%% file: root.tex
\documentclass[lettersize,journal]{IEEEtran}
\usepackage{amsmath,amsfonts}
\usepackage{algorithmic}
\usepackage{algorithm}
\usepackage{array}
\usepackage[caption=false,font=normalsize,labelfont=sf,textfont=sf]{subfig}
\usepackage{textcomp}
\usepackage{stfloats}
\usepackage{url}
\usepackage{verbatim}
\usepackage{graphicx}
\usepackage{cite}
\input{macros}

\begin{document}

\title{Hier-SLAM++: Neuro-Symbolic Semantic SLAM with a Hierarchically Categorical Gaussian Splatting}

\author{Boying Li$^{1*}$, Vuong Chi Hao$^{2}$, Peter J. Stuckey$^{1}$, Ian Reid$^{3}$, and Hamid Rezatofighi$^{1}$ 
\thanks{$^{1}$ Faculty of Information Technology, Monash University, Australia.$^{2}$ VinUniversity, Vietnam. $^{3}$ Mohamed bin Zayed University of Artificial Intelligence, United Arab Emirates. $^*$ Corresponding author: Boying Li ({\tt\small boying.li@monash.edu)}. This work is supported by the DARPA Assured Neuro Symbolic Learning and Reasoning (ANSR) program under award number FA8750-23-2-1016. The work has received partial funding from The Australian Research Council Discovery Project ARC DP2020102427.}}

\markboth{Journal of \LaTeX\ Class Files,~Vol.~14, No.~8, August~2021}%
{Shell \MakeLowercase{\textit{et al.}}: A Sample Article Using IEEEtran.cls for IEEE Journals}



\maketitle

\input{text/0_abstract}
\input{text/1_intro}
\input{text/2_related}

\input{text/3_method}

\input{text/4_experiments}

\input{text/5_conclusion}

\bibliographystyle{unsrt}
\bibliography{references}
\vfill

\section{Biography Section}
 




\begin{IEEEbiographynophoto}{Boying Li}
is a Research Fellow in the Faculty of Information Technology at Monash University, Australia. She received her Ph.D. degree in Information and Communication Engineering from Shanghai Jiao Tong University, Shanghai, China, in 2023, where she was recognized as an Outstanding Graduate. Her research interests include visual SLAM, 3D computer vision, and intelligent robotics.
\end{IEEEbiographynophoto}


\begin{IEEEbiographynophoto}{Vuong Chi Hao}
is an undergraduate student majoring in computer science at VinUniversity, Vietnam, and a research intern at the Visual and Learning for Autonomous AI Lab at Monash University. His research interests include 3D perception and long-horizon planning.
\end{IEEEbiographynophoto}


\begin{IEEEbiographynophoto}{Peter J. Stuckey}
is a Professor in the Faculty of Information Technology at Monash University and a project leader at the Data61 CSIRO laboratory. He was recognized as an ACM Distinguished Scientist in 2009 and was awarded the Google Australia Eureka Prize for Innovation in Computer Science in 2010 for his work on lazy clause generation. In 2019, he was elected a Fellow of the Association for the Advancement of Artificial Intelligence. 
Professor Stuckey is a pioneer in constraint programming, the science of modeling and solving complex combinatorial problems. His research interests include discrete optimization, programming languages, constraint-solving algorithms, bioinformatics, and constraint-based graphics.
\end{IEEEbiographynophoto}


\begin{IEEEbiographynophoto}{Ian Reid}
Professor Reid is a leading researcher in computer vision and has served as Department Chair at the Mohamed bin Zayed University of Artificial Intelligence (MBZUAI) in the United Arab Emirates. Prior to joining MBZUAI, he was Head of School and Professor of Computer Science at the University of Adelaide, Australia. A Fellow of the Australian Academy of Technological Sciences and Engineering (ATSE) and the Australian Academy of Science (AAS), and a former Rhodes Scholar, Professor Reid held an Australian Laureate Fellowship from 2013–2018 in recognition of his contributions to building Australia’s internationally competitive research capacity. He also served as Deputy Director of the Australian Centre for Robotic Vision from 2014–2021 and, in 2022, received the Australian Computer Society’s Artificial Intelligence Distinguished Researcher Award for his outstanding contributions to research in computer vision and machine learning. Over his 30-year career, Professor Reid has made significant contributions across a wide range of computer vision areas, including active vision, visual SLAM, visual geometry, human motion capture, and visual surveillance. His current research interests converge under the themes of spatial AI and embodied AI, aiming to endow real-world agents with lifelong visual learning capabilities, common-sense reasoning, and spatial awareness through video understanding for real-time robotic decision-making.
\end{IEEEbiographynophoto}


\begin{IEEEbiographynophoto}{Hamid Rezatofighi}
Dr. Hamid Rezatofighi is an Associate Professor in the Faculty of Information Technology at Monash University, Australia. His research focuses on computer vision, machine learning, and robotics, with notable contributions to robot visual perception in dynamic environments. He was previously an Endeavour Research Fellow at Stanford’s Vision and Learning Lab (SVL) and a Senior Research Fellow at the Australian Institute for Machine Learning (AIML) at the University of Adelaide. Dr. Rezatofighi completed his Ph.D. at the Australian National University in 2015. He has served as an area chair for major conferences such as CVPR, NeurIPS, ECCV, ICCV, IJCAI, and IROS, and as an associate editor for leading journals including IEEE Transactions on Image Processing and the Artificial Intelligence Journal. Over the past four years, he has secured significant research funding, including multiple DARPA projects and an ARC Discovery Project, supporting his ongoing work in computer vision and robotics.
\end{IEEEbiographynophoto}

\vfill

\newpage

\input{text/6_appendix}

\end{document}

%% file: macros.tex
\usepackage{graphics} 
\usepackage{graphicx}
\usepackage{mathptmx} 
\usepackage{times} 
\usepackage{amsmath,bm} 
\usepackage{amsfonts,amssymb} 
\usepackage{colortbl}
\usepackage[table]{xcolor}
\usepackage{booktabs}
\usepackage{multirow}
\usepackage{arydshln}
\usepackage{tabularx}
\usepackage{threeparttable} 
\usepackage{mathrsfs}
\usepackage{pifont}
\usepackage{urwchancal}
\usepackage{url}
\usepackage[most]{tcolorbox}
\usepackage{verbatim}
\usepackage{ifthen} 
\usepackage{makecell}
\usepackage{xcolor}
\usepackage[colorlinks=true, linkcolor=blue, citecolor=blue, urlcolor=blue]{hyperref} 

\newcommand{\bestcolor}[1]{\cellcolor{red!20!orange!50}{#1}}   
\newcommand{\secondcolor}[1]{\cellcolor{orange!50!yellow!60}{#1}}
\newcommand{\thirdcolor}[1]{\cellcolor{orange!40!yellow!30}{#1}}
\newcommand{\fourthcolor}[1]{\cellcolor{orange!10!yellow!30}{#1}}
\newcommand{\colorfirst}[0]{\colorbox{red!20!orange!50}{first}}
\newcommand{\colorsecond}[0]{\colorbox{orange!50!yellow!60}{second}}
\newcommand{\colorthird}[0]{\colorbox{orange!40!yellow!30}{third}}
\newcommand{\colorfourth}[0]{\colorbox{orange!10!yellow!30}{fourth}}

\newcommand{\Fig}[1]{Fig. \ref{#1}}
\newcommand{\beq}{\begin{equation}}
\newcommand{\eeq}{\end{equation}}

\newcommand{\Tab}[1]{Tab. \ref{#1}}

\newcommand{\MethodName}[1]{Hier-SLAM++}
\newcommand{\oldMethodName}[1]{Hier-SLAM}
\newcommand{\lossone}[1]{Intra-}
\newcommand{\losstwo}[1]{Inter-}

\newtcolorbox[list inside=prompt,auto counter,number within=section]{prompt}[1][]{
    colbacktitle=black!60,
    coltitle=white,
    fontupper=\footnotesize,
    boxsep=5pt,
    left=0pt,
    right=0pt,
    top=0pt,
    bottom=0pt,
    boxrule=1pt,
    title={#1},
    #1, 
}

%% file: text/0_abstract.tex
\begin{abstract}
We propose \MethodName{}, a comprehensive Neuro-Symbolic semantic 3D Gaussian Splatting SLAM method with both RGB-D and monocular input featuring an advanced hierarchical categorical representation, which enables accurate pose estimation as well as global 3D semantic mapping. 
The parameter usage in semantic SLAM systems increases significantly with the growing complexity of the environment, making scene understanding particularly challenging and costly.
To address this problem, we introduce a novel hierarchical representation that encodes both semantic and geometric information in a compact form into 3D Gaussian Splatting, leveraging the capabilities of large language models (LLMs) as well as the 3D generative model. By utilizing the proposed hierarchical tree structure, semantic information is symbolically represented and learned in an end-to-end manner.
We further introduce an advanced semantic loss designed to optimize hierarchical semantic information through both \lossone-level and \losstwo-level optimizations.
Additionally, we propose an improved SLAM system to support both RGB-D and monocular inputs using a feed-forward model. To the best of our knowledge, this is the first semantic monocular Gaussian Splatting SLAM system, significantly reducing sensor requirements for 3D semantic understanding and broadening the applicability of semantic Gaussian SLAM system.  
We conduct experiments on both synthetic and real-world datasets, demonstrating superior or on-par performance with state-of-the-art methods, while significantly reducing storage and training time requirements. 
Our project page is available at: \url{https://hierslampp.github.io/}.
\end{abstract}

\begin{IEEEkeywords}
Semantic SLAM, hierarchical category, gaussian splatting, RGB-D, monocular.
\end{IEEEkeywords}

%% file: text/1_intro.tex
\section{Introduction}

\IEEEPARstart{V}{isual} Simultaneous Localization and Mapping (SLAM) is a key technology for ego-motion estimation and scene perception, and is widely employed across various domains such as autonomous drones~\cite{heng2014autonomous}, self-driving vehicles~\cite{lategahn2011visual}, and interactive experiences in Augmented Reality (AR) and Virtual Reality (VR)~\cite{chekhlov2007ninja}.
Semantic information, which provides high-level knowledge about the environment, is indispensable for comprehensive scene understanding. It plays a crucial role in enabling intelligent robots to perform complex tasks. In recent decades, advances in image segmentation and enriched map representations have significantly accelerated progress in semantic visual SLAM~\cite{civera2011towards, bowman2017probabilistic, chang2021kimera, li2023dns, rosinol2020kimera, li2023textslam, li2020textslam, zhu2023sni, haghighi2023neural}. 

\begin{figure}[t!]		
    \centering  
     \includegraphics[width=0.57\textwidth, trim=0mm 40mm 70mm 0mm, clip]{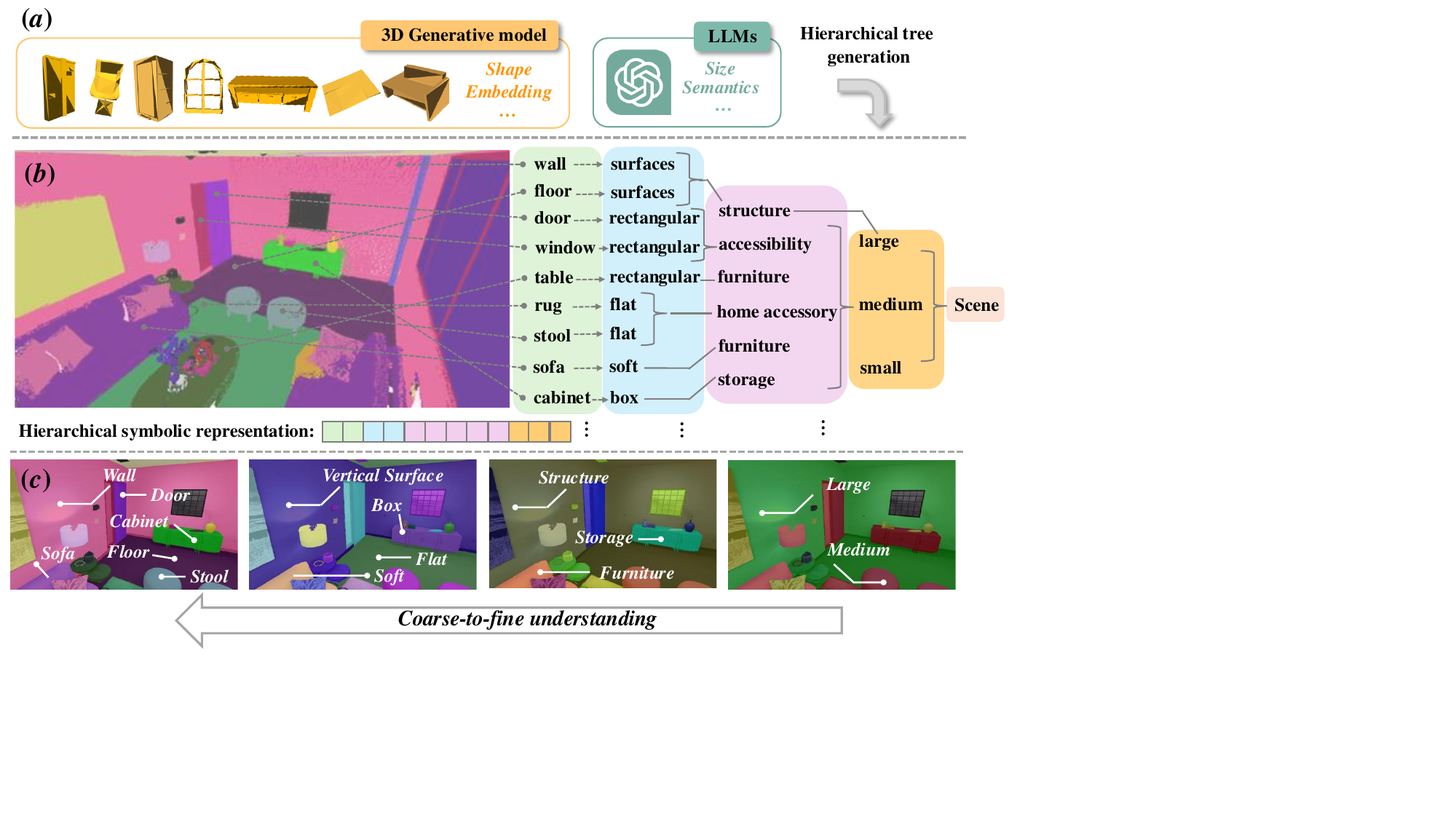} \\
    \caption{\textbf{(a).} The hierarchical tree is generated by integrating geometric information and semantic messages, utilizing 3D generative models and Large Language Models (LLMs).
    \textbf{(b).} The global 3D Gaussian map generated by \MethodName{} with learned semantic labels is shown on the left. The established hierarchical tree of the semantic information is organized on the right.
    Based on the established tree, the hierarchical symbolic representation for semantic information is shown at the bottom of the block, which compresses semantic data, reducing both memory usage and training time of the semantic SLAM.
    \textbf{(c). }The rendered semantic map at different levels shows a coarse-to-fine understanding, beneficial for real-world scenarios with shifting perspectives from distant to close.}  
    \label{fig:hierslampp}  
    \vspace{-15pt}
\end{figure}

Recently, 3D Gaussian Splatting has emerged as a powerful representation for 3D scenes~\cite{kerbl20233d, Yu2024mipsplat, Wu20244dgs}, owing to its fast rendering and optimization efficiency enabled by the highly parallelized rasterization of 3D primitives. Specifically, it models the \textit{continuous} distribution of geometric attributes using Gaussian functions. This formulation not only enhances the 3D representation fidelity but also facilitates efficient optimization, making it particularly suitable for SLAM tasks. 
While promising, current SLAM systems based on 3D Gaussian Splatting~\cite{keetha2023splatam, matsuki2024gaussian, yan2024gs, huang2024photo, yugay2023gaussian} primarily focus on geometric reconstruction. However, the absence of semantic integration limits their applicability in complex downstream tasks that require high-level scene understanding. 

To enable semantic understanding in SLAM systems,
a straightforward approach, analogous to the geometric modeling of 3D Gaussians, is to augment each 3D primitive with a discrete semantic label and represent its distribution using a categorical distribution, typically implemented via a flat Softmax-based representation.
However, 3D Gaussian Splatting is already a storage-intensive approach~\cite{lu2024scaffold, chen2024hac}, requiring numerous primitives with rich parameters to achieve high-quality rendering. 
Adding semantic distributions for each primitive further increases memory and computational costs, with complexity growing linearly with the number of semantic classes, making it impractical for complex semantic understanding.
Moreover, given a limited number of observations (e.g., video frames), introducing high-dimensional semantic parameters leads to the curse of dimensionality, hindering optimization performance.
This added overhead is especially problematic in robotics, where computational efficiency and resource constraints are critical.
Therefore, a key challenge is to design semantic encoding strategies that are both efficient and scalable, enabling robust semantic SLAM without compromising runtime and memory efficiency.

Semantic information often exhibits a natural \textit{hierarchical structure}.  
As illustrated in \Fig{fig:hierslampp}, semantic classes can be abstracted into multi-level concepts, such as \texttt{structures} and \texttt{surfaces}, which can be further organized into a hierarchical symbolic tree. 
With careful design, this structure enables rich semantics to be encoded efficiently using a small number of symbolic nodes, yielding a compact and scalable representation.
For instance, a binary tree of depth $10$ can represent up to $2^{10}$ classes using only $20$ nodes (i.e., $2 \times 10$ nodes through 2-dimensional Softmax at each level), reducing the complexity from $O(N)$ to $O(\log N)$.

Building upon this concept, we propose \MethodName{}, a comprehensive neural-symbolic semantic SLAM framework based on 3D Gaussian Splatting, featuring a novel hierarchical representation for semantic information.
To design an effective hierarchical structure, we incorporate both semantic and geometric cues, including shape and size information. Specifically, we leverage a text-to-3D generative model~\cite{siddiqui2024meshgpt} to extract shape-aware features for each class and organize them into hierarchical levels. Additionally, we employ Large Language Models (LLMs) to enrich the hierarchy with semantic relations and size priors, compensating for the limitations of generative models.
With the constructed tree, semantic information in the environment is represented as symbolic nodes along a hierarchical path and learned end-to-end during SLAM optimization.
Specifically, we propose two types of hierarchical semantic representations, one-hot and binary semantic embedding, which offer different degrees of compactness.
To enhance semantic understanding, we introduce a novel hierarchical loss function that performs joint optimization at multiple levels, including \lossone-level and \losstwo-level supervision.  
Furthermore, \MethodName{} supports both RGB-D and monocular settings. We employ the geometric output of a 3D feed-forward model~\cite{wang2024dust3r} as a depth prior, effectively removing the reliance on dedicated depth sensors. This enables monocular semantic SLAM and broadens the applicability of our system.
We further refine the 3D Gaussian Splatting-based SLAM pipeline to improve both tracking performance and runtime efficiency.  
Experiments on both synthetic and real-world datasets demonstrate that \MethodName{} achieves superior or on-par performance with state-of-the-art methods in localization, mapping, and semantic understanding under the RGB-D setting, while maintaining competitive performance in the monocular setting.
The main contributions are as follows:

1) 
We propose a hierarchical tree structure that integrates semantic, shape, and size information, constructed using LLMs and 3D generative models.  
This representation efficiently encodes semantic messages while preserving its structural hierarchy, leading to reduced memory consumption and faster training.

2)
Based on the generated hierarchical tree, we propose two types of semantic embeddings that offer different degrees of compactness.
Furthermore, we introduce a novel optimization strategy for hierarchical semantic representations, incorporating both \lossone-level and \losstwo-level losses to enable comprehensive refinement of both embeddings. 

3) 
We develop an enhanced SLAM framework that supports both RGB-D and monocular settings by leveraging a 3D feed-forward model as a geometric prior.  
To the best of our knowledge, \MethodName{} is the first Gaussian Splatting-based semantic SLAM system capable of operating with monocular input, significantly lowering sensor requirements and improving deployment flexibility.

This paper builds upon our previous work~\cite{li2024hi}, with the following major extensions:
1) The tree construction process is enhanced by integrating 3D generative models and improving the use of LLMs, resulting in a more effective symbolic tree structure.
2) In addition to the one-hot semantic representation from our earlier work, we propose a binary semantic representation for further compactness. 
3) We improve semantic optimization with a new hierarchical loss formulation and introduce a comprehensive hierarchical optimization strategy applicable to both one-hot and binary representations.
4) We propose a monocular pipeline to enable robust monocular SLAM by leveraging 3D feed-forward models, thereby extending the RGB-D semantic SLAM system into a unified framework that supports both RGB-D and monocular inputs.
Additionally, we include new real-world datasets and experiments to further validate the proposed approach.

%% file: text/2_related.tex
\section{Related Work}
\subsection{Visual SLAM system}

    Visual Simultaneous Localization and Mapping (SLAM) is a long-standing topic in computer vision and robotics. It jointly estimates camera poses and reconstructs global maps by optimizing appearance-based objectives~\cite{davison2007monoslam, klein2007parallel, mur2015orb, mur2017orb, campos2021orb, engel2017direct, forster2014svo, forster2016svo, newcombe2011dtam, newcombe2011kinectfusion, dai2017bundlefusion}.  
    One classical monocular-based approach aims to estimate accurate camera poses and a sparse global map, by minimizing the reprojection error terms \cite{mur2015orb, mur2017orb, campos2021orb}, or by minimizing the photometric losses \cite{engel2017direct, forster2014svo, forster2016svo}. 
    DTAM~\cite{newcombe2011dtam} takes a different direction by performing dense reconstruction and camera tracking with RGB inputs, though real-time capability relys on GPU support.  
    Another category of methods adopts RGB-D input for dense 3D reconstruction, which is able to provide richer information.  
    The works \cite{newcombe2011kinectfusion, dai2017bundlefusion} jointly estimate camera poses and dense surfaces from RGB-D input using a volumetric TSDF map. 
    Alongside advances in purely geometric SLAM, semantic SLAM has gained increasing attention for its ability to provide high-level scene understanding~\cite{salas2013slam++, galvez2016real, yang2019cubeslam, yang2019monocular, dong2017visual, nicholson2018quadricslam, sualeh2019simultaneous, zhi2019scenecode, rosinol2020kimera, chang2021kimera}.
    Sparse semantic SLAM methods rely on pre-scanned models~\cite{salas2013slam++, galvez2016real} or use bounding boxes~\cite{yang2019cubeslam, yang2019monocular, dong2017visual} and quadrics~\cite{nicholson2018quadricslam}, though accuracy remains limited.  
    In contrast, semantic understanding in dense SLAM becomes more tractable and effective when combined with dense observations from RGB-D inputs~\cite{mccormac2017semanticfusion, runz2018maskfusion, grinvald2019volumetric} or dense visual SLAM pipelines~\cite{zhi2019scenecode}. 
    More recently, new paradigms have emerged to enhance dense SLAM with richer scene understanding across both monocular and RGB-D inputs, such as implicit neural fields and 3D Gaussian Splatting.
    Furthermore, semantic integration allows current SLAM systems to move beyond geometric reconstruction to deliver high-level scene interpretation, such as class-level mapping, semantic navigation, and task-aware perception. This significantly broadens its applications in real-world fields.

    Some works, such as Kimera~\cite{Rosinol20rss-dynamicSceneGraphs}, adopt dynamic scene graphs for representing \textit{entities} within the observed environment ~\cite{kim20193, armeni20193d, wald2020learning, wu2021scenegraphfusion, zhang2025open}.
    In contrast, we propose a generalizable hierarchical symbolic representation for modeling \textit{semantics} in the world, leveraging large language models and 3D generative priors, along with symbolic capabilities~\cite{besold2021neural, wang2024imperative} to abstract semantic concepts without relying on scene-specific definitions.
    Crucially, our symbolic hierarchy is carefully designed to produce compact semantic embeddings and is seamlessly integrated into the neural learning process within the SLAM system, offering both structural abstraction and interpretability for semantic encoding.

\subsection{Neural implicit visual SLAM}

    Neural Radiance Fields (NeRF) \cite{mildenhall2021nerf} introduce a novel scene representation paradigm, inspiring the integration of NeRF with SLAM pipelines for simultaneous localization and high-quality mapping. Unlike traditional SLAM methods that rely on explicit geometry, NeRF-SLAM models represent scenes implicitly, enabling continuous and photorealistic reconstruction from sequential views.
    Early work such as iNeRF~\cite{yen2021inerf} explored solely pose tracking with a fixed pre-trained neural field.
    iMap~\cite{sucar2021imap} advanced this idea by proposing a keyframe-based RGB-D SLAM that jointly optimizes camera poses and a global implicit map using a single MLP.
    NICE-SLAM~\cite{zhu2022nice} enhances robustness and scalability by introducing a hierarchical neural implicit representation, which is later extended to the monocular setting in NICER-SLAM~\cite{zhu2024nicer}.
    Recent works focus on improving efficiency: ESLAM~\cite{johari2023eslam} leverages a hybrid representation combining implicit TSDFs with multi-scale feature planes to achieve faster operation; Point-SLAM~\cite{sandstrom2023point} adopts a point-based neural representation to enable efficient and detail-preserving dense SLAM.
    To further enhance scene understanding beyond geometry and appearance, researchers also incorporate semantics into NeRF-based SLAM systems \cite{li2023dns, zhu2023sni, haghighi2023neural, zhai2024nis}. 
    Some works leverage the semantic information extracted from 2D observations to integrate them into the visual SLAM system and generate the global semantic map \cite{li2023dns, zhai2024nis}.
    SNI-SLAM \cite{zhu2023sni} proposes a semantic RGB-D SLAM which learn semantics by integrating geometry, appearance, and semantic features into a shared feature space.
    However, these methods remain constrained by the slow convergence and high overhead of neural implicit representations~\cite{Hu2022CVPR, Liu2023ICCV}, especially when jointly optimizing semantics.
    The inherent inefficiency of NeRF representations still conflicts with the real-time demands of SLAM systems—let alone the added cost from semantic integration.
    In contrast, Gaussian Splatting provides significant advantages, including fast rendering performance and high-density reconstruction quality, making it a promising new kind of map representation.

\subsection{3D Gaussian Splatting SLAM}

    3D Gaussian Splatting has recently emerged as a promising 3D representation due to its fast rendering and high-fidelity reconstruction. Compared to implicit representations, it models the environment as a collection of explicit Gaussian distributions with geometric and appearance attributes. Several SLAM systems have adopted this representation in their pipelines.
    SplaTAM~\cite{keetha2023splatam} proposes a RGB-D SLAM framework with isotropic Gaussians representation using silhouette guidance for pose estimation, while MonoGS~\cite{matsuki2024gaussian} presents an efficient solution supporting both monocular and RGB-D inputs.
    Photo-SLAM~\cite{huang2024photo} uses 3D points from ORB-SLAM to initialize Gaussian positions; GS-SLAM~\cite{yan2024gs} introduces a coarse-to-fine tracking strategy with sparse Gaussian selection; and Gaussian-SLAM~\cite{yugay2023gaussian} utilizes DROID-SLAM poses to manage multiple 3D Gaussian submaps. These approaches showcase the versatility of 3D Gaussian Splatting in various SLAM settings.
    However, integrating semantics into Gaussian Splatting-based SLAM remains challenging. Jointly optimizing geometry, appearance, and semantics involves heterogeneous value ranges and significantly increases memory and computational demands.
    GS$^3$LAM~\cite{li2024gs3lam} fuses 2D semantic labels into an RGB-D SLAM framework, with embeddings sized according to the number of semantic classes.
    SGS-SLAM~\cite{li2024sgs} appends RGB 3-channels to support semantic visualization. SemGauss-SLAM~\cite{zhu2024semgauss} leverages flat semantic embeddings supervised by foundation models, though this introduces high computational cost and overlooks the hierarchical structure of real-world semantics.
    To address these limitations, we propose a compact and efficient solution that fully exploits the hierarchical nature of semantic categories. Our method encodes semantic information into a hierarchical symbolic tree, generated through the synergy of large language models and 3D generative priors, and learns this semantic representation end-to-end within the SLAM framework operation.

%% file: text/3_method.tex
\section{Method}

\begin{figure*}[t!]		
    \centering  
    \includegraphics[width=1.0\textwidth, trim=0mm 105mm 40mm 0mm, clip]{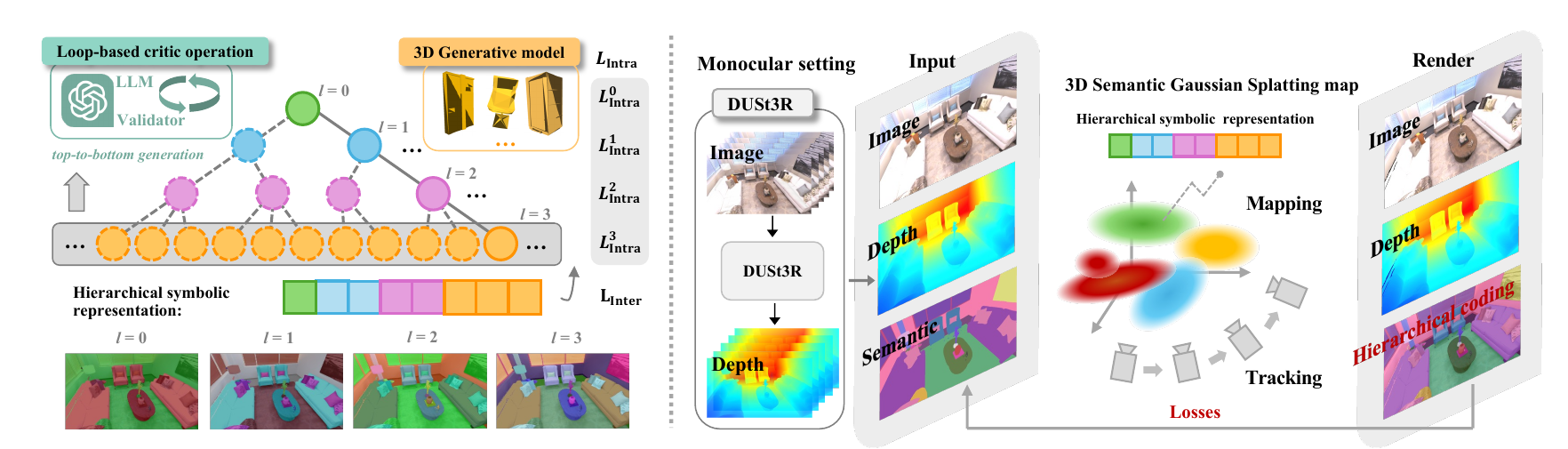} \\
    \caption{Overview of the \MethodName{} pipeline. \textbf{Top Left:} Hierarchical representation of semantic information. The Tree Generation process leverages the capabilities of both LLMs and 3D Generative Models (see Section III-A). This hierarchical tree is used to establish a symbolic coding for each Gaussian primitive. Additionally, we introduce a novel loss function that combines Intra-level Loss \( L_\text{Intra} \) and Inter-level Loss \( L_\text{Inter} \) to optimize the hierarchical semantic representation (see Section III-B).  
    \textbf{Bottom Left:} An example of hierarchical semantic rendering. 
    \textbf{Right:} The global 3D Gaussian map is initialized using the first frame of the video stream input. The system then alternates between the \textit{Tracking} and \textit{Mapping} steps as new frames are processed (see Section III-D). In the RGB-D setting, depth is directly obtained from sensor input, whereas in the monocular setting the 3D feed-forward method (DUSt3R) is used to generate a geometric prior for depth estimation (see Section III-C). } 
    \label{fig:method_pipeline}
\end{figure*}

The overall pipeline of our method is illustrated in \Fig{fig:method_pipeline}.
At the core of our approach, we model the entire semantic space as a compact hierarchical structure, where each semantic class is represented by a symbolic hierarchy.
To construct a generative hierarchical tree that captures both semantic relations and associated geometric attributes, we leverage Large Language Models (LLMs) and 3D generative models. The hierarchical representation and tree generation process are detailed in Section III-A.
Based on the proposed structure, semantic information is encoded into the embedding space in two formats: a \textit{one-hot representation} and a \textit{binary representation}, both of which are hierarchically organized and compact.
To support end-to-end training, we introduce hierarchical optimization with two loss functions, \lossone_level and \losstwo_level, which jointly optimize the semantic space based on incoming observations.
The two proposed tree encoding schemes and hierarchical optimization are discussed in detail in Section III-B.
In the monocular setting, we leverage geometric priors extracted from a feed-forward 3D model to guide learning during SLAM, incorporating online depth adjustment and geometric supervisory signals, as described in Section III-C.
Finally, the full semantic Gaussian Splatting SLAM system is presented in Section III-D.

\subsection{Hierarchical Representation and Generation}
Leveraging the inherent hierarchy of semantic classes, we propose a hierarchical symbolic representation to encode semantic information in a compact and efficient manner, enhancing the scalability of semantic SLAM system. 
This section presents a detailed description of the proposed hierarchical tree parametrization and its generation process using both LLMs and 3D generative models.

\subsubsection{Hierarchical Tree Parametrization} 
The proposed hierarchical representation is abstracted as a multi-level tree, formulated as $\bm{G} = (\bm{V}, \bm{E})$, where $\bm{V}$ and $\bm{E}$ denote the sets of vertices and edges, respectively, as illustrated in \Fig{fig:tree_rep}.
The vertex set is defined as $\bm{V} =\left\{v_{l}\right\}_{l=0}^{L}$, where $\bm{v}_{l}$ represents the set of semantic classes at level $l$ of the tree hierarchy, spanning from the root level ($l=0$) to the deepest level ($l=L$). 
The edge set $\bm{E} = \left\{e_{m}\right\}_{m=0}^{L-1}$ encodes the subordinate relationships between adjacent levels in the hierarchy, capturing semantic attribution, relative scale, and geometric structure.
The original semantic classes are represented as leaf nodes located at the deepest level ($l = L$) of the hierarchical tree. For the $i$-th semantic class $g^{i}$, its hierarchical representation is defined as:
\beq
g^{i} = \{ v_{l}^{i}, e_{m}^{i} \mid l = 0,1,...,L; \; m = 0,1,...,L-1 \},
\label{eq:single_class_tree}
\eeq
which corresponds to a hierarchical path for the semantic meaning: $ g^{i} = v_{0}^{i} \xrightarrow{e_{0}} v_{1}^{i} \xrightarrow{e_{1}} \cdots \xrightarrow{e_{L-2}} v_{L-1}^{i} \xrightarrow{e_{L-1}} v_{L}^{i}$.
For example, the 5-level hierarchical symbolic representation of the class $\texttt{Bed}$ can be expressed as:
$g^{\texttt{bed}} = \left\{v_{0}:\texttt{LargeItems}\right\} \rightarrow \left\{v_{1}: \texttt{Furnishings}\right\} \rightarrow \left\{v_{2}: \texttt{BedroomFurniture}\right\} \rightarrow \left\{v_{3}: \texttt{Rectangular}\right\} \rightarrow \left\{v_{4}: \texttt{Bed}\right\}$.
In this representation, the original semantic label $g^i$ is abstracted as a hierarchical path from a coarse category to a fine-grained concept. At each level, symbolic nodes $v_{l}^{i}$ represent the semantic attributes of the class, while inter-level relationships, such as inclusion and possession, are described by the edge elements $e_{m}^{i}$.
As will be demonstrated in Section III-B, with a carefully structured design, the proposed hierarchical representation yields a compact and computationally efficient representation. 
It allows each semantic class to be expressed in a progressive, interpretable, and symbolic manner, incorporating both semantic abstraction and geometric structure. 
Moreover, from a hierarchical perspective, the original flat representation can be regarded as a degenerate case of a single-level tree, enabling a unified formulation.
For clarity and notational simplicity, we omit the superscripts (i.e., class indices) of variables in the remainder of this section.

\subsubsection{Tree Generation via LLMs and 3D Generative Models}

To represent semantic information in a reasonable, comprehensive, and compact manner, the tree structure should integrate both semantic and geometric cues, two fundamental sources of information in unknown environments that are essential for achieving a holistic understanding of the global 3D scene.
To extract high-level semantic concepts, including functional relationships and abstract descriptors, we employ large language models (LLMs), specifically GPT-4 Turbo \cite{gpt4o_turbo2024}, which is well known for its strong commonsense reasoning and language understanding abilities.
On the other hand, to capture general geometric information beyond dataset-specific constraints, we employ 3D generative models, particularly text-to-3D frameworks. These models are capable of producing diverse and geometrically accurate 3D objects from textual prompts after appropriate training, offering a unified and generalizable geometric representation. 
Specifically, we adopt MeshGPT \cite{siddiqui2024meshgpt}, a text-to-3D model that learns symbolic 3D embeddings for the local mesh geometry and topology of each object. This property makes it particularly well-suited for our goal of constructing geometry-aware hierarchical semantic structures.
However, the text-to-3D method generates 3D objects within a unified coordinate system but omits explicit size information.
To address this limitation, we complement the generated geometric shapes with size attributes inferred from large language models (LLMs), enabling a more comprehensive geometric representation for each semantic class.
Leveraging both large language models (LLMs) and 3D generative models enables the construction of an elaborate and well-structured hierarchical tree, where each level captures a distinct aspect of a class’s attributes—semantic, functional, or geometric, which further enriches the depth and expressiveness of global 3D scene understanding.
The remainder of this section presents a detailed description of the hierarchical tree generation process.

\begin{figure}[!t]		
    \centering  
    \includegraphics[width=0.9\textwidth, trim=0mm 85mm 50mm 0mm, clip]{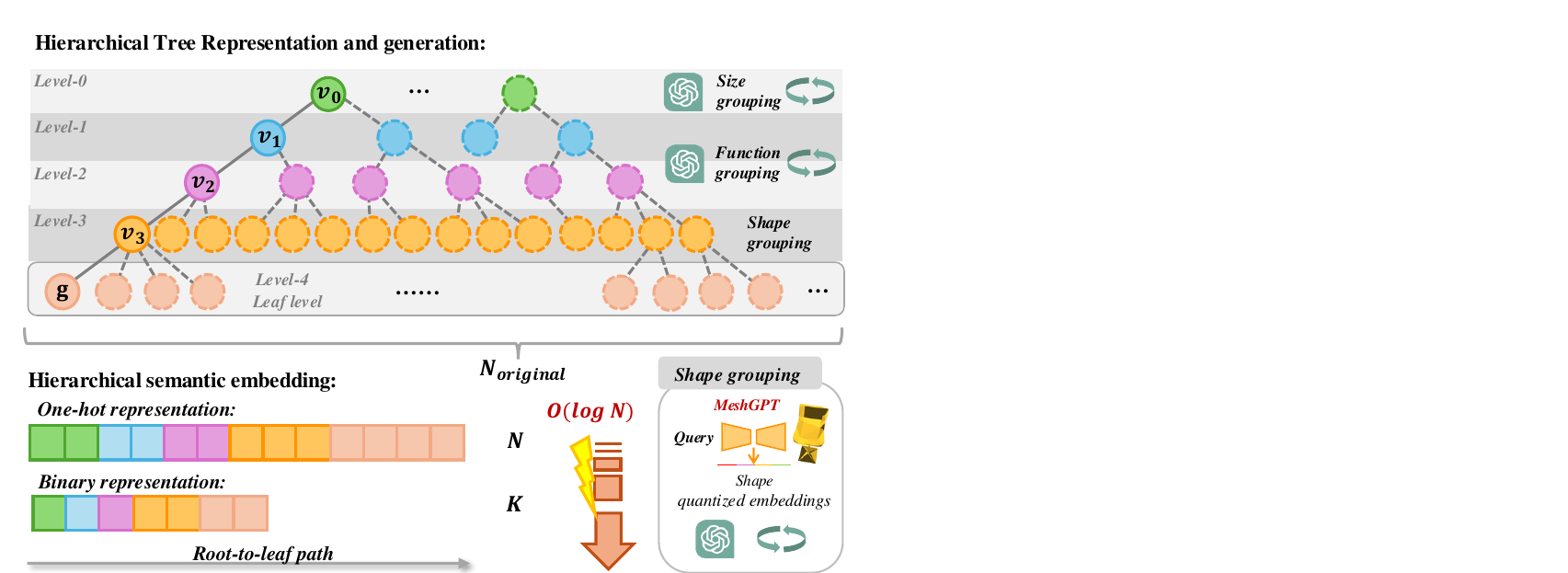} \\
    \caption{Visualization of the hierarchical tree structure and semantic embedding.
    For \textbf{tree representation}, each semantic class is expressed hierarchically as $g = \{ v_{l}, e_{m} \}$, where edges $e_{m}$ are depicted as lines connecting tree nodes. The nodes within the same level are illustrated via the same color.
    For \textbf{tree generation}, we utilize both LLMs and 3D Generative Models to extract and group messages.
    For \textbf{hierarchical semantic embedding}, we represent semantic information using a hierarchical symbolic path. We propose two types of representations: the one-hot representation and the binary representation, which can reduce the original dimensionality by up to $O(\log N)$.}
    \label{fig:tree_rep}  
\end{figure}

The overall process of hierarchical tree generation is illustrated in \Fig{fig:tree_rep}.
At the top level, the initial grouping divides all semantic classes into several subgroups based on their typical physical sizes in real-world scenes, which are obtained through prompt-based queries to the LLM. The generation of level-0 message through this size-based grouping is formulated as:
\beq
\{ v, e \}_{0} = \mathcal{L}^\text{size} \{ g \}
\eeq
\noindent where $\mathcal{L}^\text{size}$ denotes the LLM-driven grouping operator.
The resulting clusters categorize the semantic classes into groups such as \texttt{SmallItems}, \texttt{MediumItems}, and \texttt{LargeItems}.
Details of the specific prompt used to query the LLM are provided in Appendix-I.

Next, the semantic classes within each size-based group are used as input prompts to the LLM for further clustering based on semantic relationships, with an emphasis on functional properties. This step generates the following levels structure, formulated as:
\beq
\{ v, e \}_{1} = \mathcal{L}^\text{func} \{ g \}
\eeq
\noindent where $\mathcal{L}^\text{func}$ denotes the LLM-driven clustering operator based on functional similarity.
It is worth noting that $\{ v, e \}_{1}$ does not refer strictly to a single hierarchical level. Rather, it denotes a generic stage of functional grouping that follows the initial size-based partitioning.
Depending on the balance and structural design of the overall tree, the functional clustering process may be applied multiple times to generate several intermediate levels. This hierarchical refinement enables the representation to achieve maximum semantic compactness while maintaining accuracy and interpretability.
The specific prompt used for this functional clustering is provided in Appendix-I.

After generating the preceding semantic substructures using LLMs across multiple levels, 3D generative models are employed to further cluster semantic classes based on their geometric characteristics. This geometry-based symbolic representation complements the language-derived hierarchy, enabling a unified structure that incorporates both semantic and geometric information. 
Specifically, we adopt MeshGPT~\cite{siddiqui2024meshgpt} to capture geometric information for each semantic class. This text-to-3D framework encodes object shapes and internal topological relationships into latent, quantized embeddings, resulting in a symbolic 3D representation within the geometric shape space.
Given a text prompt (i.e., the class name) as input to the pre-trained MeshGPT model, we extract the latent quantized embeddings to represent the geometric information of the class:
\begin{equation}
\bm{z} = \mathcal{E}\{g\} \in \mathbb{N}^{F \times 2}
\end{equation}
\noindent where $\mathcal{E}$ denotes the pre-trained MeshGPT encoder, and $\bm{z}$ is the quantized geometric embedding extracted for class $g$. Here, $F$ represents the number of triangle faces used in the mesh representation \cite{siddiqui2024meshgpt}.
Since the number of triangle faces $F$ varies across semantic objects, we first compute an average shape embedding $ \bm{z}_{\text{avg}} \in \mathbb{R}^{1 \times 2} $ for each class. These averaged embeddings are then concatenated and normalized to a standard distribution (i.e., zero mean and unit variance) to ensure consistency across classes:
\beq
\tilde{\bm{z}} = \mathcal{N}\{ \mathcal{C}(\bm{z}_\text{avg})\}
\eeq
\noindent where $\mathcal{N}$ and $\mathcal{C}$ denote the normalization and concatenation operations, respectively.
The resulting normalized embeddings $\tilde{\bm{z}}$ are subsequently clustered using the K-means++ algorithm \cite{arthur2006k}, denoted by the operator $\mathcal{K}$. 
To address the lack of descriptive labels for the resulting clusters, we further employ an LLM-based summarization module to interpret and refine the clustering outcomes:
\beq
\{ v, e \}_{2} = \mathcal{L}^\text{sum} \{ \mathcal{K}(\tilde{\bm{z}}) \}
\eeq
\noindent where $\mathcal{L}^\text{sum}$ denotes the use of an LLM to generate semantic summaries for each geometric cluster.
The resulting descriptive labels, such as \textit{“box”, “flat”, or “soft”}, capture the dominant geometric attributes within each group, thereby enhancing the interpretability of the hierarchical structure. Details of the LLM prompts and summarization process are provided in Appendix-I.

Whenever LLMs are employed, we ensure that the outputs are balanced, meaningful, and descriptively coherent through carefully designed prompt inputs, as detailed in Appendix-I.
To further ensure the structural correctness within each level of the hierarchical tree, we introduce a \textit{loop-based critic operation} performs multiple iterations.
In each iteration, an LLM is followed by a validator module that checks the quality of the generated clusters.
Specifically, within each LLM-based layer, the validator assesses the clustering results in terms of completeness and semantic consistency by comparing them against the input prompt.
Correctly grouped nodes are retained, while erroneously included classes are removed.
Any omitted classes are compiled into a new prompt and fed into the next iteration of LLM-based clustering.
This process repeats until no omitted classes remain at the current level, ensuring a complete and semantically accurate hierarchical structure.
Finally, a comprehensive validation process is applied to the entire hierarchical structure, including an automated check to detect and resolve duplicates or omissions across levels, followed by a manual inspection to ensure correctness and conceptual consistency. This ensures a fully balanced and semantically coherent tree.

\subsection{Hierarchical Semantic Coding and Optimization}

Based on the generated hierarchical tree structure, semantic information is encoded as compact, symbolic hierarchical codes and integrated into the 3D Gaussian primitives, which are learned in an end-to-end fashion from the input image stream.
To ensure accurate and structured semantic understanding, we introduce a semantic hierarchical loss, composed of both \lossone-level and \losstwo-level components. These losses are incorporated into the overall optimization during the online SLAM process. 
The following sections provide a detailed description of the hierarchical semantic encoding and the proposed optimization framework.

\subsubsection{Hierarchical Semantic Coding} 
As introduced in Section III-A, we model the semantic space as a hierarchical tree structure. Within this hierarchical space, each semantic class is represented in a structured and compact form that preserves essential semantic information.
The proposed hierarchical semantic coding enhances the joint optimization of the SLAM system and offers valuable scalability for semantic understanding in the complex environments.
Specifically, we introduce two types of hierarchical semantic coding strategies, both of which compress the original flat semantic labels while retaining crucial semantic messages.

\textbf{One-hot representation:} 
As illustrated in \Fig{fig:tree_rep}, the hierarchical semantic embedding $\bm{h}$ is constructed by concatenating one-hot embeddings from all levels of the tree:
\beq
\bm{h} = \mathcal{C}(\bm{h}_{l}) \in \mathbb{B}^N, \quad \bm{h}_{l} \in \mathbb{B}^{n}
\eeq
\noindent where subscript denotes the $l$-th tree level, and $\mathcal{C}$ represents the concatenation operator.
At each level, $\bm{h}_l$ is an $n$-dimensional one-hot vector (as shown in colored boxes in \Fig{fig:tree_rep}) used to represent symbolic nodes (colored circles) within that level. The dimension $n$ corresponds to the maximum number of nodes at level $l$, indexing node representations from $0$ to $n-1$. The total dimensionality of the hierarchical semantic embedding is given by $N = \sum n$, i.e., the sum over all levels.
Compared to the original flat representation of dimension $N_{\text{original}}$, this hierarchical structure enables an information-preserving reduction of up to $O(\log N)$ in representational complexity.
The resulting compact semantic code significantly reduces memory usage and facilitates efficient joint optimization in the SLAM system.

\textbf{Binary representation:} 
To further improve compactness, we propose a binary version of the hierarchical semantic representation. The binary embedding $\bm{b}$ is constructed by concatenating the binary vectors $\bm{b}_l$ across all levels of the hierarchical tree:
\beq
\bm{b} = \mathcal{C}(\bm{b}_{l}) \in \mathbb{B}^K, \quad \bm{b}_{l} \in \mathbb{B}^{k}, \quad \text{where } k = \left\lceil \log_2 n \right\rceil
\eeq
\noindent where $\bm{b}_l$ is the binary code converted from the one-hot embedding $\bm{h}_l$ at level $l$, i.e., $\bm{b}_l = \text{Binary}(\bm{h}_l)$, as illustrated in the final row of \Fig{fig:tree_rep}.
For example, if the one-hot vector has dimension $n = 8$ (representing $8$ distinct semantic nodes indexed from $0$ to $7$), its binary representation requires only $k = 3$ bits for $2^3 = 8$.
More generally, the binary code length at each level is reduced to $k = \lceil \log_2 n \rceil$, compared to the $n$-dimensional one-hot encoding.

\subsubsection{Loss Calculation} 
To effectively and efficiently optimize the proposed hierarchical semantic codings, we introduce a hierarchical loss formulation that supervises semantic embedding both \textit{within each level (intra-level)} and \textit{across levels (inter-level)}.
Specifically, the total semantic loss is defined as:
\beq
L_\text{{Semantic}} = \omega_1 L_\text{{Intra}} + \omega_2 L_\text{{Inter}}
\eeq
\noindent where \( \omega_1 \) and \( \omega_2 \) are balancing coefficients for each loss term.
Specifically, the intra-level loss $L_\text{Intra}$ is employed within each level's embedding $\bm{h}^{l}$ to ensure the optimization of semantic information within each level:
\beq
L_\text{{Intra}} = \sum_{l=0}^{L} L_\text{{ce}}(\mathcal{S}(\bm{h}^{l}), \mathcal{P}^{l})
\eeq
\noindent where $L_\text{{ce}}$ indicates the cross-entropy semantic loss, and $\mathcal{P}^{l}$ denotes the semantic ground truth at level $l$. $\mathcal{S}$ stands for the Softmax operation, converting embeddings into probabilities. 
For our proposed compact version with binary representation $\bm{b}^l$, we employ the binary cross-entropy loss $L_\text{bce}$, instead of the standard cross-entropy, for computing the binary intra-level loss:
\beq
L_\text{Inter}^\text{bin} = \sum_{l=0}^{L} L_\text{{bce}}(\mathcal{S}(\bm{b}^{l}), \mathcal{P}^{l})
\eeq

To ensure accurate semantic understanding across all levels of the hierarchical structure, we introduce the inter-level loss $L_{\text{Inter}}$, which enforces consistency between the hierarchical representation and the original flat semantic distribution.
To this end, we employ a lightweight decoder to transform the holistic hierarchical embedding $\bm{h}$ into flat semantic predictions.
Specifically, the decoder consists of two stacked 2D convolutional layers, with a ReLU activation applied between them. The output is then passed through a softmax function to produce a probability distribution over the flat semantic classes:
\beq
\text{decoder}(\bm{h}) = \text{Conv} \big(\text{ReLU}(\text{Conv}(\bm{h}))\big)
\eeq
The inter-level loss is defined as:
\beq
L_{\text{Inter}} = L_{\text{ce}}\big(\mathcal{S}(\text{decoder}(\bm{h})), \mathcal{P}\big),
\eeq where $\mathcal{S}(\cdot)$ denotes the softmax function, $\mathcal{P}$ is the ground truth flat semantic label, and $L_{\text{CE}}$ is the standard cross-entropy loss.

\subsection{Monocular setting with geometric priority}

Accurate depth information provides valuable geometric cues for global map reconstruction and pose estimation in SLAM systems.  
While RGB-D methods directly acquire depth from sensors, monocular SLAM approaches~\cite{mur2015orb, mur2017orb, campos2021orb, engel2017direct} must rely on multi-view triangulation to infer depth and initialize mapping, which increases the difficulty of simultaneously performing localization and mapping.
To address this limitation, a number of methods~\cite{tateno2017cnn, luo2018real, laidlow2019deepfusion, ye2020drm, tiwari2020pseudo, zhang2023hi} introduce depth priors from learned depth networks into SLAM systems, achieving promising results.  
More recently, a new trend has emerged: \textit{3D feed-forward models}, which can generate accurate global 3D representations of previously unseen environments after training.
In our work, we adopt a precise and highly generalizable 3D feed-forward method to obtain the geometric prior used in monocular SLAM.  
Specifically, we employ DUSt3R~\cite{wang2024dust3r}, which produces both 3D point clouds and pose estimations from sparse-view image inputs, serving as a strong geometric prior in our system.
Details of the monocular SLAM setting in \MethodName{} are provided in the following section.

\subsubsection{Geometric prior acquisition}

DUSt3R \cite{wang2024dust3r} supports sparse view inputs with a minimum of two views. Larger view inputs enhance reconstruction accuracy by providing more information, yet at the cost of increased computational demands. 
To balance reconstruction accuracy and computational efficiency, we sample every \( S \) frames with a skip of \( \delta \) frames from the input video to form an image set \( \{I_i\} \). 
\beq
\{I_i \; | \; i = i_{0} + \delta \cdot (s - 1), \; s = 1, \dots, S\}
\eeq 

Each image set is then fed into DUSt3R to generate geometric information, specifically the depth maps.  
While DUSt3R is also capable of estimating camera poses for each input frame, our experiments reveal that its pose predictions exhibit relatively low accuracy, often performing worse than a simple constant-velocity pose model when integrated into a SLAM pipeline. Therefore, we retain only the geometric (depth) information predicted by DUSt3R, as it provides more reliable guidance for our SLAM system.

\subsubsection{Geometric prior correction}
The depth predicted by DUSt3R is scale-ambiguous and lacks metric consistency across image sets.
This leads to differences in scale between the predicted geometry and the SLAM system’s estimated global map, which evolves over time.
We therefore incorporate a geometric prior correction step to resolve this scale inconsistency.

To align the scale of the depth prior with the current global map, we first estimate the scale and shift parameters by comparing the rendered depth from the SLAM map with the predicted depth from DUSt3R.
Specifically, for each input frame $I_c$, we render a depth map $D_c$ and silhouette map $S_c$ from the global 3D Gaussian map $G$ using the current camera pose $\mathbf{T}_c$:
\beq
\{ D_c, S_c \} = \mathcal{R}(G, \mathbf{T}_c)
\eeq
\noindent where $\mathcal{R}$ represents the rasterization procedure for the 3D Gaussian map. 
We then compute the global scale and shift parameters $\lambda_c$ and $\tau_c$ by aligning the DUSt3R-predicted depth $\hat{D}c$ to the rendered depth $D_c$ using least squares estimation over the visible regions $M_c$, as defined by the silhouette map $S_c$:
\beq
\{ \lambda_c, \tau_c \} = \arg\min \left( \hat{D}_c^{\text{align}} - D_c \right)^2 M_c
\eeq
\noindent where $\hat{D}_c^{\text{align}} = \lambda_c \cdot \hat{D}_c + \tau_c$ denotes the aligned depth.
This simple affine transformation is sufficient to resolve the scale ambiguity inherent in the geometric priors obtained from DUSt3R, while preserving their relative 3D structure for integration into the monocular SLAM pipeline.

Secondly, the calculated scale and shift parameters are used as initial values and are jointly optimized with the 3D Gaussians during the map optimization process, guided by a depth supervision loss:
\beq
\{ \lambda_c, \tau_c, G \} = \arg\min \big( | \hat{D}_c^{\text{align}} - D_c | M_c + L_\text{reg} \big)
\eeq
\noindent where $L_\text{reg}$ is a regularization term defined as the absolute deviation between the optimized parameters and their initialization, preventing overfitting or convergence to poor local minima. 
By incorporating this optimization into the SLAM operation, the depth prior can be continuously refined online as new observations arrive, effectively providing geometric supervision throughout the SLAM process.
This not only improves the quality of the global map reconstruction but also enhances the accuracy of camera pose estimation.

\subsection{Semantic Gaussian Splatting SLAM}

The global semantic 3D Gaussian Splatting SLAM system is illustrated in \Fig{fig:method_pipeline}.
Built upon the semantic 3D Gaussian representation, global reconstruction (Mapping) and camera pose estimation (Tracking) are alternately performed over the input video stream.
In the RGB-D setting, depth information for each frame is directly acquired from the depth sensor.
In the monocular setting, a feed-forward model is employed to generate geometric priors, as described in the previous section.
The construction of the semantic 3D Gaussians and the overall semantic SLAM system are detailed in the following sections.

\subsubsection{Semantic 3D Gaussian representation} 
Our semantic 3D Gaussian representation encodes both geometric and semantic information using Gaussian primitives augmented with hierarchical semantic embeddings.
Motivated by \cite{keetha2023splatam}, we adopt isotropic, view-independent Gaussians and integrate them with semantic feature vectors.
Each semantic Gaussian is parameterized by its color $\bm{c}$, center position $\bm{\mu}$, radius $r$, opacity $o$, and semantic embedding $\bm{h}$.
The influence of each Gaussian $G(\bm{X})$ \cite{kerbl20233d} at 3D point $\bm{X}$ follows the standard Gaussian formulation:
\beq
G(\bm{X}) = o \: \exp\left(-\frac{||\bm{X}-\bm{\mu}||^2}{2r^2}\right)
\eeq
\noindent where $\bm{X}$ stands for the 3D point. 

Following \cite{kerbl20233d}, every semantic 3D Gaussian primitive within the current global map is projected onto the image plane using the tile-based differentiable $\alpha$-compositing rendering. 
The semantic map $H$ is rendered as follows:
\begin{equation}
H = \sum_{i=1}^{n} \bm{h}_{i} \alpha_{i}(\bm{X}) T_{i} \quad \text{with} \quad T_{i} = \prod_{j=1}^{i-1} (1 - \alpha_{j}(\bm{X}))
\end{equation}
The rasterized color image $C$, depth image $D$, and the silhouette image $S$ are defined as follows:
\begin{align}
C = \sum_{i=1}^{n} \bm{c}_{i} \alpha_{i}(\bm{X}) T_{i}, \;
D = \sum_{i=1}^{n} \bm{d}_{i} \alpha_{i}(\bm{X}) T_{i}, \;
S = \sum_{i=1}^{n} \alpha_{i}(\bm{X}) T_{i}
\end{align}
Different from previous work \cite{keetha2023splatam}, which uses separate forward and backward Gaussian modules for different parameters, our \MethodName{} system adopts a unified forward and backward modules that processes all parameters, including semantics, color, depth, and silhouette images, which significantly improving overall runtime efficiency.

\subsubsection{Tracking}
The tracking step focuses on pose estimation for each frame.  
For each incoming frame, the constant velocity model is first applied to initialize its pose based on the previous frame's pose estimation.  
Following that, a pose optimization is performed with the global Gaussian map $G$ fixed:
\beq
\mathbf{T}_c = \arg\min_{\mathbf{T}_c} L_{\text{Track}}(G, \mathbf{T}_c)
\eeq
\noindent where the tracking loss includes the rendering color and depth losses:
\beq
L_\text{{Track}} = M \big( w_{1} L_\text{{Depth}}(G, \mathbf{T}_c) + w_{2} L_\text{{Color}}(G, \mathbf{T}_c) \big)
\eeq
\noindent where $L_\text{Depth}$ and $L_\text{Color}$ denote the L1-loss for the rendered depth and color information. The rendering follows the tile-based differentiable $\alpha$-compositing procedure, utilizing the current pose estimation along with the current global map estimation.
Here, weights $ w_{1}$ and $w_{2}$ are introduced to balance the two loss terms, and the optimization is only performed on the silhouette-visible image regions, defined as $M = (S > \delta)$.
For the RGB-D setting, depth supervisory signals are directly obtained from depth sensors, whereas for the monocular setting, we leverage the geometric prior derived from the feed-forward method.

\subsubsection{Mapping} 
The global map information, including semantic information, is optimized in the mapping procedure while keeping camera poses $\mathbf{T}_i$ fixed:
\beq
G = \arg\min_{G} L_{\text{Map}}(G, \mathbf{T}_i)
\eeq
The mapping optimization losses include depth, color, and semantic losses:
\beq
L_\text{{Map}} = w_{3}  M  L_\text{{Depth}}(G, \mathbf{T}_i) + w_{4} L_\text{{Color}}'(G, \mathbf{T}_i) + w_{5} L_\text{{Semantic}} (G, \mathbf{T}_i)
\eeq
\noindent where $L_\text{{Semantic}}$ is the proposed semantic loss introduced in Section III-B, and $L_\text{{Color}}'$ is the weighted sum of SSIM color loss and L1-Loss. The weights $w_{3}$, $w_{4}$, and $w_{5}$ are used to balance the different loss terms.
Similar to the tracking process, depth supervisory signals come from sensors in the RGB-D setting.  
For the monocular setting, geometric priors derived from the feed-forward method are applied, and the two parameters (scale and shift) are further jointly optimized within the mapping process.

%% file: text/4_experiments.tex
\section{Experiments}

\subsection{Experiment settings}

The experiments are conducted on both synthetic and real-world datasets, including ScanNet \cite{Dai2017scannet}, Replica \cite{straub2019replica}, and TUM-RGBD \cite{sturm12iros}.
Following the evaluation metrics which used in previous SLAM methods \cite{keetha2023splatam, zhu2022nice}, we leverage ATE RMSE (cm) to quantify SLAM tracking accuracy.
For mapping evaluation, Depth L1 (cm) is used to measure accuracy.
To assess image rendering quality, we adopt PSNR (dB), SSIM, and LPIPS as evaluation metrics.
In line with previous works \cite{zhu2024semgauss, li2023dns, zhu2023sni}, we assess semantic rendering performance using mIoU (mean Intersection over Union across all classes) to represent global semantic information.  
To evaluate computational efficiency, we report the runtime of the proposed SLAM method.
For comparison, we benchmark our approach against state-of-the-art dense visual SLAM methods, including both NeRF-based and 3D Gaussian SLAM approaches, to demonstrate its effectiveness. Additionally, we compare with state-of-the-art semantic SLAM methods, covering both NeRF-based and Gaussian-based techniques, to highlight its capability in hierarchical semantic representation and scalability.
Experiments are conducted on an NVIDIA L40S GPU, while runtime tests are evaluated on both the L40S and RTX 4090 to enable broader comparisons.

For experimental settings, the semantic embedding of each Gaussian primitive is initialized randomly. The semantic optimization loss weights $\omega_1$ and $\omega_2$ are set to $1.0$ and $0.0$, respectively, for the first $\eta$ iterations, where $\eta$ is set as $15$. Afterwards, $\omega_1$ remains $1.0$, while $\omega_2$ is increased to $5.0$. This strategy means that we first apply the Inter-level loss to initialize the hierarchical semantic coding, followed by incorporating the Cross-level loss to refine the whole embedding. 
For tracking loss, we use $\delta=0.99$, with weights $w_{1}=1.0$ and $w_{2}=0.5$.
For mapping, the loss weights are set as follows: $w_{3}=1.0$, $w_{4}=0.5$, and $w_{5}=0.2$.
The two types of semantic embeddings proposed in our method, one-hot and binary, are evaluated and reported across multiple experiments, including tracking, mapping, rendering, runtime, and semantic understanding, providing a comprehensive ablation study on the effectiveness and compactness of the hierarchical semantic representation.

\begin{table}[!tb]   
    \centering
    \caption{RGB-D Tracking performance ATE RMSE (cm) on the Replica. \\Best results are highlighted as  
    \colorfirst, \colorsecond, \colorthird.} 
    \renewcommand{\arraystretch}{1.1} 
    \setlength{\tabcolsep}{2pt}
    \begin{tabular}{lccccccccc}
        \hline
        \rowcolor{white}
        \toprule
        \textbf{Methods} & \textbf{Avg.} & \textbf{R0} & \textbf{R1} & \textbf{R2} & \textbf{Of0} & \textbf{Of1} & \textbf{Of2} & \textbf{Of3} & \textbf{Of4} \\
        \hline
        iMap \cite{sucar2021imap} & 4.15 & 6.33 & 3.46 & 2.65 & 3.31 & 1.42 & 7.17 & 6.32 & 2.55 \\
        NICE-SLAM \cite{zhu2022nice} & 1.07 & 0.97 & 1.31 & 1.07 & 0.88 & 1.00 & 1.06 & 1.10 & 1.13 \\  
        Vox-Fusion \cite{yang2022vox} & 3.09 & 1.37 & 4.70 & 1.47 & 8.48 & 2.04 & 2.58 & 1.11 & 2.94 \\ 
        co-SLAM \cite{wang2023co} & 1.06 & 0.72 & 0.85 & 1.02 & 0.69 & 0.56 & 2.12 & 1.62 & 0.87 \\
        ESLAM \cite{johari2023eslam} & 0.63 & 0.71 & 0.70 & 0.52 & 0.57 & 0.55 & 0.58 & 0.72 & 0.63 \\
        Point-SLAM \cite{sandstrom2023point} & 0.52 & 0.61 & \thirdcolor{0.41} & 0.37 & 0.38 & 0.48 & 0.54 & 0.69 & 0.72 \\
        {MonoGS-RGBD \cite{matsuki2024gaussian}} & 0.79 & 0.47 & 0.43 & 0.31 & 0.70 & 0.57 & 0.31 & \bestcolor{0.31} & 3.2 \\  
        SplaTAM \cite{keetha2023splatam} & \thirdcolor{0.36} & 0.31 & \secondcolor{0.40} & 0.29 & 0.47 & 0.27 & \thirdcolor{0.29} & \secondcolor{0.32} & 0.55 \\
        \hdashline
        SNI-SLAM \cite{zhu2023sni} & 0.46 & 0.50 & 0.55 & 0.45 & 0.35 & 0.41 & 0.33 & 0.62 & \secondcolor{0.50} \\
        DNS SLAM \cite{li2023dns} & 0.45 & 0.49 & 0.46 & 0.38 & \thirdcolor{0.34} & 0.35 & 0.39 & 0.62 & 0.60 \\
        SemGauss-SLAM \cite{zhu2024semgauss} & \secondcolor{0.33} & 0.26 & 0.42 & \thirdcolor{0.27} & \thirdcolor{0.34} & \thirdcolor{0.17} & 0.32 & 0.36 & \bestcolor{0.49} \\
        SGS-SLAM \cite{li2024sgs} & 0.41 & 0.46 & 0.45 & 0.29 & 0.46 & 0.23 & 0.45 & 0.42 & 0.55 \\
        
        \oldMethodName{} \cite{li2024hi} &  \secondcolor{0.33} &  \bestcolor{0.21} & 0.49 & \secondcolor{0.24} & \bestcolor{0.29} & \secondcolor{0.16} & 0.31 & 0.37 & 0.53 \\ 
        \hdashline
       
        \textbf{\MethodName{}} (one-hot) & \bestcolor{0.31} &  \thirdcolor{0.24} & \bestcolor{0.36} & \bestcolor{0.23} & \secondcolor{0.30} & \bestcolor{0.15} & \secondcolor{0.28} & 0.39 & \thirdcolor{0.51} \\  
        \textbf{\MethodName{}} (binary) & \bestcolor{0.31} & \secondcolor{0.23} & 0.46 & \bestcolor{0.23} & \bestcolor{0.29} & \bestcolor{0.15} & \bestcolor{0.27} & \thirdcolor{0.34} & 0.54 \\ 
         \toprule
    \end{tabular}
    \label{tab:exp_pose_replica}
\end{table}

\subsection{Camera Tracking Accuracy}
\Tab{tab:exp_pose_replica} presents our tracking performance on the Replica dataset \cite{straub2019replica} in the RGB-D setting, compared with state-of-the-art dense SLAM methods, both with and without semantic information.  
By accurately learning semantic information, our \MethodName{} outperforms all existing methods in 6 out of 8 sequences.
For the remaining two sequences, our method achieves nearly the same performance with sub-millimeter level localization error. The lowest average tracking error clearly demonstrate that our approach surpasses other state-of-the-art methods.
We present our RGB-D tracking performance on the real-world ScanNet dataset \cite{Dai2017scannet} in \Tab{tab:exp_pose_scannet}. 
Compared to the results on the Replica dataset, all methods exhibit degraded performance due to the noisy and sparse depth sensor input, as well as lower color image quality caused by motion blur.  
Specifically for semantic understanding, the semantic annotations provided by ScanNet are significantly noisier than those in Replica, containing noticeable noise and imprecise boundaries. All of these issues negatively impact the overall performance of semantic SLAM systems.  
We evaluate all six sequences and show that our method achieves performance comparable to state-of-the-art approaches \cite{li2024hi, keetha2023splatam}, demonstrating the robustness of our proposed method under these challenging conditions.
The tracking performance on the TUM-RGBD dataset is shown in \Tab{tab:exp_pose_tumrgbd}. As this dataset does not provide semantic ground truth, we evaluate our method solely based on tracking accuracy.
Without access to semantic information, our system operates as a standard visual SLAM pipeline.
As a result, the tracking accuracy becomes comparable to the state-of-the-art method \cite{keetha2023splatam} in the TUM-RGBD dataset.

\begin{table}[!tb]
    \centering
    \setlength{\tabcolsep}{2pt}
    \caption{RGB-D Tracking performance ATE RMSE (cm) on the Scannet. \\ Best results are highlighted as 
    \colorfirst, \colorsecond, \colorthird, \colorfourth.
    }
    \label{tab:exp_pose_scannet}
    \begin{tabular}{lccccccc}
        \toprule
        \textbf{Methods} & \textbf{Avg.} & \textbf{0000} & \textbf{0059} & \textbf{0106} & \textbf{0169} & \textbf{0181} & \textbf{0207}  \\
        \midrule
        NICE-SLAM \cite{zhu2022nice} & \secondcolor{10.70} & 12.00 & 14.00 & \bestcolor{7.90} & \thirdcolor{10.90} & 13.40 & \secondcolor{6.20}  \\
        Vox-Fusion \cite{yang2022vox} & 26.90 & 68.84 & 24.18 & \secondcolor{8.41} & 27.28 & 23.30 & 9.41  \\
        Point-SLAM \cite{sandstrom2023point} & 12.19 & \bestcolor{10.24} & \bestcolor{7.81} & \thirdcolor{8.65} & 22.16 & 14.77 & 9.54  \\
        SplaTAM \cite{keetha2023splatam} & 11.88 & 12.83 & 10.10 &17.72 & 12.08 & 11.10 & 7.46  \\
        SGS-SLAM \cite{li2024sgs} & \bestcolor{9.87} & \secondcolor{11.15} & \thirdcolor{9.54} & \fourthcolor{10.43} & \secondcolor{10.70} & 11.28 & \bestcolor{6.11} \\
        SemGauss-SLAM \cite{zhu2024semgauss} & -- & \fourthcolor{11.87} & \secondcolor{7.97} & -- & \bestcolor{8.70} & \bestcolor{9.78} & 8.97  \\
        \oldMethodName{} \cite{li2024hi} & \thirdcolor{11.36} & \thirdcolor{11.45} & 9.61 & 17.80 & 11.93 & \thirdcolor{10.04} & \fourthcolor{7.32} \\
        \hdashline
        \textbf{\MethodName{}} (one-hot) & 11.72 & 12.93 & \fourthcolor{9.59} & 17.70 & \fourthcolor{11.63} & \fourthcolor{10.98} & 7.51 \\  
        \textbf{\MethodName{}} (binary) & \fourthcolor{11.59} & 13.24 & 9.60 & 17.83 & 11.75 & \secondcolor{9.83} & \thirdcolor{7.29} \\
        \bottomrule 
    \end{tabular}
\end{table}

\begin{table}[!tb]
    \centering
    \setlength{\tabcolsep}{1.2pt}
    \caption{RGB-D Tracking performance ATE RMSE (cm) on the TUM-RGBD. Best results are highlighted as  
    \colorfirst, \colorsecond, \colorthird, \colorfourth.}
    \label{tab:exp_pose_tumrgbd}
    \begin{tabular}{lcccccc}
        \toprule
        \textbf{Methods} & \textbf{Avg.} & \textbf{fr1/desk} & \textbf{fr1/desk2} & \textbf{fr1/room} & \textbf{fr2/xyz} & \textbf{fr3/off}  \\
        \midrule
        Kintinuous \cite{whelan2015real} & \secondcolor{4.84} & 3.70 & 7.10 & \secondcolor{7.50} & 2.90 & \thirdcolor{3.00} \\
        ElasticFusion \cite{whelan2015elasticfusion} & 6.91 & \secondcolor{2.53} & 6.83 & 21.49 & \secondcolor{1.17} & \secondcolor{2.52} \\
        ORB-SLAM3 \cite{campos2021orb} & \bestcolor{2.21} & \bestcolor{1.60} & \bestcolor{2.37} & \bestcolor{5.83} & \bestcolor{0.32} & \bestcolor{0.94} \\
        NICE-SLAM \cite{zhu2022nice} & 15.87 & 4.26 & \thirdcolor{4.99} & 34.49 & 31.73 & 3.87 \\
        Vox-Fusion \cite{yang2022vox} & 11.31 & 3.52 & 6.00 & 19.53 & 1.49 & 26.01 \\
        Point-SLAM \cite{sandstrom2023point} & 8.92 & 4.34 & \secondcolor{4.54} & 30.92 & \fourthcolor{1.31} & \fourthcolor{3.48} \\
        SplaTAM \cite{keetha2023splatam} & \thirdcolor{5.48} & \fourthcolor{3.35} & \fourthcolor{6.54} & \thirdcolor{11.13} & \thirdcolor{1.24} & 5.16 \\
        \hdashline
        \textbf{\MethodName{}} & \fourthcolor{5.62} & \thirdcolor{3.31} & 6.59 & \fourthcolor{11.73} & 1.34 & 5.14  \\ 
        \bottomrule 
    \end{tabular}
\end{table}

\begin{table}[!t]
    \centering
    \caption{Reconstruction performance Depth L1 (cm) on Replica. Best results are highlighted as \colorfirst, \colorsecond, \colorthird.}
    \renewcommand{\arraystretch}{1.1} 
    \setlength{\tabcolsep}{2.0pt} 
    \begin{tabular}{lccccccccc}
        \hline
        \toprule
        \textbf{Methods} & \textbf{Avg.} & \textbf{R0} & \textbf{R1} & \textbf{R2} & \textbf{Of0} & \textbf{Of1} & \textbf{Of2} & \textbf{Of3} & \textbf{Of4} \\
        \hline
        NICE-SLAM \cite{zhu2022nice} & 2.97 & 1.81 & 1.44 & 2.04 & 1.39 & 1.76 & 8.33 & 4.99 & 2.01 \\
        Vox-Fusion \cite{yang2022vox} & 2.46 & 1.09 & 1.90 & 2.21 & 2.32 & 3.40 & 4.19 & 2.96 & 1.61 \\
        Co-SLAM \cite{wang2023co} & 1.51 & 1.05 & 0.85 & 2.37 & 1.24 & 1.48 & 1.86 & 1.66 & 1.54 \\
        ESLAM \cite{johari2023eslam} & 0.95 & 0.73 & 0.74 & 1.26 & 0.71 & 1.02 & 0.93 & 1.03 & 1.18 \\
        
        SNI-SLAM \cite{zhu2023sni} & 0.77 & \secondcolor{0.55} & 0.58 & 0.87 & \thirdcolor{0.55} & 0.97 & 0.89 & \bestcolor{0.75} & 0.97 \\
        SGS-SLAM \cite{li2024sgs} & \bestcolor{0.36} & -- & -- & -- & -- & -- & -- & -- & -- \\
        SemGauss-SLAM \cite{zhu2024semgauss} & 0.50 & \bestcolor{0.54} & \thirdcolor{0.46} & \thirdcolor{0.43} & \bestcolor{0.29} & 0.22 & \bestcolor{0.51} & 0.98 & \secondcolor{0.56} \\
        \oldMethodName{} \cite{li2024hi} & \thirdcolor{0.49} & 0.58 & \bestcolor{0.40} & \secondcolor{0.40} & \bestcolor{0.29} & \secondcolor{0.19} & \bestcolor{0.51} & \thirdcolor{0.95} & \thirdcolor{0.57} \\
        \hdashline
        \textbf{\MethodName{}} (one-hot) & \secondcolor{0.48} & \thirdcolor{0.56} & \secondcolor{0.41} & \bestcolor{0.39} & \bestcolor{0.29} & \thirdcolor{0.20} & \thirdcolor{0.53} & \secondcolor{0.92} & \bestcolor{0.55} \\ 
        \textbf{\MethodName{}} (binary) & \thirdcolor{0.49} & 0.59 & \bestcolor{0.40} & \secondcolor{0.40} & \secondcolor{0.30} & \bestcolor{0.17} & \secondcolor{0.52} & \thirdcolor{0.95} & \thirdcolor{0.57} \\    
        \toprule
    \end{tabular}
    \label{tab:exp_depth}
\end{table}

\subsection{Global Mapping Performance}
Following prior work \cite{li2024hi, li2024sgs, zhu2024semgauss}, we evaluate depth accuracy on the Replica dataset \cite{straub2019replica}, which provides high-quality ground truth depth maps for reliable evaluation of overall mapping performance.
The quantitative results are presented in \Tab{tab:exp_depth}.
Our method achieves the lowest depth error in 7 out of 8 sequences and maintains a consistently low average error across all sequences, with only a $0.1$ cm difference compared to the best-performing baseline.
By leveraging semantic understanding, our approach enables the learning of precise geometry in unseen environments, which in turn benefits both tracking and mapping within the whole semantic SLAM system.
These results confirm the strong semantic mapping capability of our method and highlight its superior performance in dense 3D reconstruction.

\subsection{Rendering Quality}
Rendering performance is becoming an increasingly important evaluation metric, particularly for SLAM systems built upon new scene representations, such as neural implicit representations and gaussian splatting.
Following prior works \cite{sandstrom2023point, zhu2022nice}, the evaluation is conducted on the input views of the Replica dataset.
We present a quantitative analysis of the rendering performance of our proposed method in \Tab{tab:exp_render_replica}.
Our method outperforms state-of-the-art approaches in rendering quality across all 8 sequences, as well as in the overall average, surpassing both semantic and non-semantic SLAM baselines.
The accurate semantic understanding learned by our system enhances both geometric and appearance representations through joint optimization, further improving the overall rendering fidelity of the SLAM system.

\begin{figure*}[!t]		
    \centering  
    \includegraphics[width=1.1\textwidth, trim=60mm 75mm 40mm 0mm, clip]{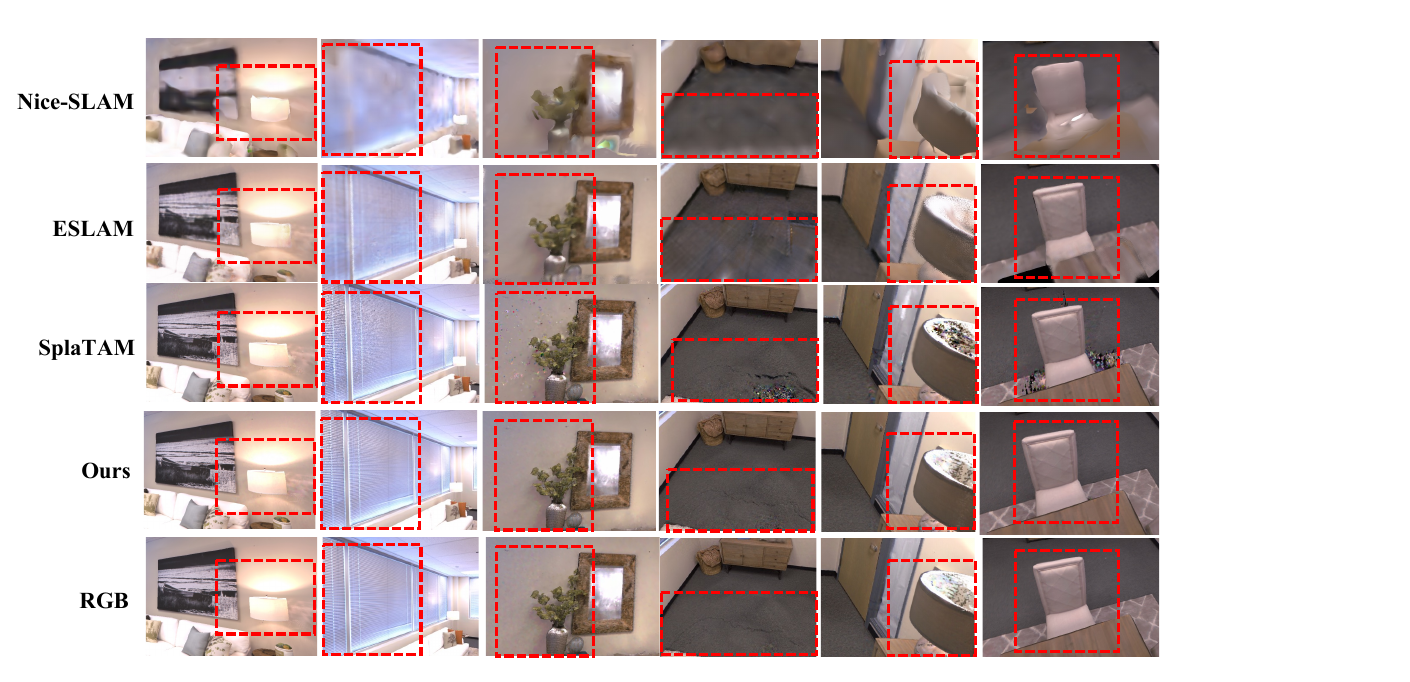} \\
    \caption{Visualization of RGB image rendering performances on the Replica dataset \cite{straub2019replica}.  
    \textbf{The first to third rows} show the rendering outputs of the comparison methods. 
    \textbf{The fourth row} presents the image rendering results of our \MethodName{}.  
    \textbf{The last row} displays the ground truth images for reference.  
    We use red boxes to highlight key differences regions within the image. The visual comparison clearly demonstrates that our method achieves superior rendering quality, closely resembling the ground truth images compared to state-of-the-art approaches.}  
    \label{fig:exp_render} 
\end{figure*}

The visual comparison of rendered images is shown in \Fig{fig:exp_render}.
Benefiting from accurate semantic understanding and an improved SLAM framework, our method achieves significantly better detail preservation.
Notable improvements include sharper and more accurate lamp boundaries in the first and fifth columns, as well as fine structural details, such as the shutters in the second column.
Furthermore, our method produces high-quality renderings with fewer noise artifacts, evident in the clearer floor in the fourth column and the cleaner reconstruction of the chair in the sixth column.
These results demonstrate that improved mapping quality directly contributes to enhanced rendering fidelity.
The joint optimization of semantic and photometric losses proves mutually beneficial: semantic learning facilitates more accurate global 3D reconstruction, which in turn leads to better rendering performance in our \MethodName{} system.

\begin{table*}[!ht]
    \centering
    \caption{Rendering performance PSNR, SSIM, LPIPS on Replica. Best results are highlighted as \colorfirst, \colorsecond, \colorthird.}
    \label{tab:exp_render_replica}
    \begin{tabular}{llcccccccccccc}
        \toprule
        \textbf{Methods} & \textbf{Metrics} & \textbf{Avg.} & \textbf{room0} & \textbf{room1} & \textbf{room2} & \textbf{office0} & \textbf{office1} & \textbf{office2} & \textbf{office3} & \textbf{office4} \\
        \midrule
        \multicolumn{11}{c}{\textbf{Visual SLAM}} \\
        \multirow{3}{*}{NICE-SLAM \cite{zhu2022nice}} 
        & PSNR $\uparrow$ & 24.42 & 22.12 & 22.47 & 24.52 & 29.07 & 30.34 & 19.66 & 22.23 & 24.94 \\
        & SSIM $\uparrow$ & 0.809 & 0.689 & 0.757 & 0.814 & 0.874 & 0.886 & 0.797 & 0.801 & 0.856 \\
        & LPIPS $\downarrow$ & 0.233 & 0.330 & 0.271 & 0.208 & 0.229 & 0.181 & 0.235 & 0.209 & 0.198  \\
        \hline
        \multirow{3}{*}{Vox-Fusion \cite{yang2022vox}} 
        & PSNR $\uparrow$ & 24.41 & 22.39 & 22.36 & 23.92 & 27.79 & 29.83 & 20.33 & 23.47 & 25.21 \\
        & SSIM $\uparrow$ & 0.801 & 0.683 & 0.751 & 0.798 & 0.857 & 0.876 & 0.794 & 0.803 & 0.847 \\
        & LPIPS $\downarrow$ & 0.236 & 0.303 & 0.269 & 0.234 & 0.241 & 0.184 & 0.243 & 0.213 & 0.199 \\
        \hline
        \multirow{3}{*}{Co-SLAM \cite{wang2023co}} 
        & PSNR $\uparrow$ & 30.24 & 27.27 & 28.45 & 29.06 & 34.14 & 34.87 & 28.43 & 28.76 & 30.91  \\
        & SSIM $\uparrow$ & 0.939 & 0.910 & 0.909 & 0.932 & 0.961 & 0.969 & 0.938 & 0.941 & 0.955 \\
        & LPIPS $\downarrow$ & 0.252 & 0.324 & 0.294 & 0.266 & 0.209 & 0.196 & 0.258 & 0.229 & 0.236 \\
        \hline
        \multirow{3}{*}{ESLAM \cite{johari2023eslam}} 
        & PSNR $\uparrow$ & 29.08 & 25.32 & 27.77 & 29.08 & 33.71 & 30.20 & 28.09 & 28.77 & 29.71 \\
        & SSIM $\uparrow$ & 0.929 & 0.875 & 0.902 & 0.932 & 0.960 & 0.923 & 0.943 & 0.948 & 0.945 \\
        & LPIPS $\downarrow$ & 0.239 & 0.313 & 0.298 & 0.248 & 0.184 & 0.228 & 0.241 & 0.196 & 0.204 \\
        \hline
        \multirow{3}{*}{SplaTAM \cite{keetha2023splatam}} 
        & PSNR $\uparrow$ & 34.11 & 32.86 & 33.89 & 35.25 & 38.26 & 39.17 & 31.97 & 29.70 & 31.81 \\
        & SSIM $\uparrow$ & 0.968 & \secondcolor{0.978} & 0.969 & 0.979 & 0.977 & 0.978 & 0.969 & 0.949 & 0.949 \\
        & LPIPS $\downarrow$ & 0.102 & 0.072 & 0.103 & 0.081 & 0.092 & 0.093 & 0.102 & 0.121 & 0.152 \\
        \midrule
        \multicolumn{11}{c}{\textbf{Semantic SLAM}} \\
        \multirow{3}{*}{SNI-SLAM \cite{zhu2023sni}} 
        & PSNR $\uparrow$ & 29.43 & 25.91 & 28.17 & 29.15 & 31.85 & 30.34 & 29.13 & 28.75 & 30.97 \\
        & SSIM $\uparrow$ & 0.921 & 0.884 & 0.900 & 0.921 & 0.935 & 0.925 & 0.930 & 0.932 & 0.936 \\
        & LPIPS $\downarrow$ & 0.237 & 0.307 & 0.292 & 0.265 & 0.185 & 0.211 & 0.230 & 0.209 & 0.198 \\
        \hline
        \multirow{3}{*}{SGS-SLAM \cite{li2024sgs}} 
        & PSNR $\uparrow$ & 34.66 & 32.50 & 34.25 & 35.10 & 38.54 & 39.20 & 32.90 & 32.05 & 32.75 \\
        & SSIM $\uparrow$ & 0.973 & \thirdcolor{0.976} & \secondcolor{0.978} & \thirdcolor{0.981} & 0.984 & 0.980 & 0.967 & \thirdcolor{0.966} & 0.949 \\
        & LPIPS $\downarrow$ & 0.096 & 0.070 & 0.094 & 0.070 & 0.086 & 0.087 & 0.101 & 0.115 & 0.148 \\
        \hline
        \multirow{3}{*}{SemGauss-SLAM \cite{zhu2024semgauss}} 
        & PSNR $\uparrow$ & 35.03 & 32.55 & 33.32 & 35.15 & 38.39 & 39.07 & 32.11 & 31.60 & 35.00 \\
        & SSIM $\uparrow$ & \bestcolor{0.982} & \bestcolor{0.979} & \thirdcolor{0.970} & \secondcolor{0.987} & \bestcolor{0.989} & 0.972 & \bestcolor{0.978} & \bestcolor{0.972} & \bestcolor{0.978} \\
        & LPIPS $\downarrow$ & \bestcolor{0.062} & \bestcolor{0.055} & \bestcolor{0.054} & \bestcolor{0.045} & \bestcolor{0.048} & \secondcolor{0.046} & \bestcolor{0.069} & \bestcolor{0.078} & \bestcolor{0.093} \\
        \hline
        
        \multirow{3}{*} { \oldMethodName{} \cite{li2024hi}}
        & PSNR $\uparrow$ & \bestcolor{35.70} & \bestcolor{32.83} & \bestcolor{34.68} & \secondcolor{36.33} & \bestcolor{39.75} & \bestcolor{40.93} & \thirdcolor{33.29} & \secondcolor{32.48} & \thirdcolor{35.33} \\ 
        & SSIM $\uparrow$ & \secondcolor{0.980} & \thirdcolor{0.976} & \bestcolor{0.979} & \secondcolor{0.987} & \secondcolor{0.988} & \secondcolor{0.989} & \thirdcolor{0.975} & \secondcolor{0.971} & 0.976 \\
        & LPIPS $\downarrow$ & \thirdcolor{0.067} & \thirdcolor{0.060} & 0.063 & 0.052 & \secondcolor{0.050} & \thirdcolor{0.049} & 0.083 & \secondcolor{0.081} & \secondcolor{0.094} \\

        \hline
        
        \multirow{3}{*} { \textbf{\MethodName{}} (one-hot)}
        & PSNR $\uparrow$ & \thirdcolor{35.61} & \secondcolor{32.76} & \secondcolor{34.64} & \thirdcolor{36.27} & \secondcolor{39.65} & \thirdcolor{40.12} & \secondcolor{33.46} & \bestcolor{32.49} & \bestcolor{35.52} \\ 
        & SSIM $\uparrow$ & \thirdcolor{0.980} & \secondcolor{0.978} & \bestcolor{0.979} & \secondcolor{0.987} & \thirdcolor{0.987} & \thirdcolor{0.985} & \secondcolor{0.977} & \secondcolor{0.971} & \secondcolor{0.977} \\ 
        & LPIPS $\downarrow$ & \thirdcolor{0.067} & \secondcolor{0.058} & \secondcolor{0.058} & \thirdcolor{0.051} & \thirdcolor{0.053} & 0.064 & \thirdcolor{0.077} & \thirdcolor{0.082} & 0.100 \\ 

        \hdashline
        \multirow{3}{*} { \textbf{\MethodName{}} (binary)}
        & PSNR $\uparrow$ & \secondcolor{35.68} & \secondcolor{32.34} & \thirdcolor{34.67} & \bestcolor{36.47} & \thirdcolor{39.61} & \secondcolor{40.82} & \bestcolor{33.59} & \thirdcolor{32.46} & \secondcolor{35.47} \\ 
        & SSIM $\uparrow$ & \secondcolor{0.981} & 0.975 & \bestcolor{0.979} & \bestcolor{0.988} & \thirdcolor{0.987} & \bestcolor{0.990} & \bestcolor{0.978} & \secondcolor{0.971} & \thirdcolor{0.976} \\ 
        & LPIPS $\downarrow$ & \secondcolor{0.066} & 0.064 & \thirdcolor{0.061} & \secondcolor{0.049} & 0.056 & \bestcolor{0.043} & \secondcolor{0.076} & \secondcolor{0.081} & \thirdcolor{0.098} \\ 
        \bottomrule
    \end{tabular}
\end{table*}

\subsection{Running time}
The runtime results for the compared methods are provided in \Tab{tab:runtime}, covering both NeRF-based and Gaussian Splatting-based SLAM methods. 
For the reason that one compared baseline using a flat representation (\MethodName{}-flat) cannot run successfully on an NVIDIA RTX 4090 due to GPU memory limitations, we report the runtime separately for RTX 4090 and L40S.
In the first block of \Tab{tab:runtime}, we observe that, compared to NeRF-based methods \cite{zhu2022nice}, Gaussian-based state-of-the-art methods \cite{keetha2023splatam} achieve significantly higher operating speeds, benefiting from the fast rendering and efficient optimization capabilities inherent to Gaussian Splatting representation. 
Furthermore, with the carefully refined forward and backward gaussian module introduced in our system, which serves as the core component of gaussian splatting-based SLAM systems, our method (\MethodName{}- w/o sem) achieves up to 2.4× faster tracking and 2.2× faster mapping compared to the best-performing method \cite{keetha2023splatam}.  
In the second block of \Tab{tab:runtime}, we evaluate the impact of different semantic representations on runtime performance.
Our method maintains high efficiency by employing hierarchical semantic coding (\MethodName{}-one-hot), achieving nearly 3× faster tracking compared to the flat semantic representation (\MethodName{}-flat).
Furthermore, our proposed binary semantic representation (\MethodName{}-binary) demonstrates even greater efficiency, enabling more lightweight performance.
The flat version is slightly faster in mapping, with an improvement of 0.38 seconds per-frame, this is due to its cross-entropy loss calculation directly without hierarchical optimization for the proposed methods.
Notably, our \MethodName{} achieves a rendering speed of 2000 FPS, which increases to 3000 FPS when semantic information is not included.

\begin{table}[!tb]
\centering
\caption{Runtime on Replica/R0. Best results are highlighted as \textbf{first}.}
\setlength{\tabcolsep}{1.5pt}
\label{tab:runtime}
\begin{tabular}{c|lcccc}
\hline
\toprule 
 & \multirow{2}{*}{Methods} & Tracking & Mapping & Tracking & Mapping \\
 & & /Iteration & /Iteration & /Frame & /Frame \\
\hline
\multirow{5}{*}{\textbf{RTX 4090}} & NICE-SLAM \cite{zhu2022nice} & 122.42 & 104.25 & 1.22 & 6.26 \\ 
& SplaTAM \cite{keetha2023splatam} & 44.27 & 50.07 & 1.77 & 3.00 \\ 
& \textbf{\MethodName{}} (w/o sem) & \textbf{18.71} & \textbf{22.93} & \textbf{0.75} & \textbf{1.38} \\ 
& \textbf{\MethodName{}} (one-hot)  & 37.63 & 194.78 & 1.50 & 11.69 \\
& \textbf{\MethodName{}} (binary)  & 31.22 & 92.63 & 1.25 & 5.56 \\
\hline
\multirow{4}{*}{\textbf{L40S}} &  \oldMethodName{} \cite{li2024hi} & 61.23 & \textbf{170.30} & 2.45 & \textbf{10.22} \\   
& \textbf{\MethodName{}} (flat) & 168.94 & 204.25 & 6.75 & 12.26 \\
& \textbf{\MethodName{}} (one-hot) & 62.21 & 212.62 & 2.49 & 12.76 \\
& \textbf{\MethodName{}} (binary) & \textbf{58.91} & 210.65 & \textbf{2.36} & 12.64 \\

\hline
\end{tabular}
\begin{tablenotes} 
            \footnotesize
            \itshape
            \raggedright 
            \item \textbf{\MethodName{} (w/o sem)} represents our proposed system without semantic information.
            \item The \textbf{units} are as follows: Tracking/Iteration (ms), Mapping/Iteration (ms), Tracking/Frame (s), and Mapping/Frame (s).
        \end{tablenotes}
\vspace{-10pt}
\end{table}

\subsection{Hierarchical semantic understanding}
We perform semantic understanding experiments on the synthetic dataset Replica \cite{straub2019replica} to demonstrate the comprehensive performance of our proposed method. 
Replica \cite{straub2019replica} is a synthetic indoor dataset that includes 102 semantic classes with high-quality semantic annotations. 
Similar to methods \cite{zhu2024semgauss, li2023dns, zhu2023sni}, we present our quantitative results, evaluated in mIoU ($\%$) across all original semantic classes (102 classes) in \Tab{tab:replica_semantic_segmentation}, where the rendered semantic map is compared against the semantic ground truth. 
Additionally, consistent with the previous approach \cite{li2024sgs}, we also report mIoU scores on a subset of semantic classes in the second block of the table. 
To further analyze the storage requirements of semantic SLAM methods, we report the parameter storage usage in the second block of \Tab{tab:replica_semantic_segmentation}.
The generated hierarchical tree used in our method is shown in Appendix-II, where semantic concepts are organized into symbolic nodes, as illustrated by the colored nodes.  

From \Tab{tab:replica_semantic_segmentation}, our method, along with the flat coding version and the proposed one-hot version, both demonstrate superior semantic performance compared to all existing methods. 
By utilizing hierarchical semantic representation, \MethodName{} (one-hot) achieves a similar semantic performance with less than a $1\%$ difference in mIoU while requiring 4.5× less storage compared to the flat version. 
The training time shown in \Tab{tab:runtime} demonstrates that our hierarchical version (\MethodName{} (one-hot)) requires only 37\% of the tracking time per frame while maintaining similar frame mapping performance compared to the flat version (\MethodName{} (flat)). The comparable mIoU performance, along with significantly reduced storage and computational cost, clearly demonstrates the effectiveness and efficiency of our proposed method.
Furthermore, our binary version (\MethodName{} (binary)) achieves state-of-the-art semantic understanding while achieves the best parameter storage efficiency, reducing storage usage to 63\% of the one-hot version and requiring only 20\% of the storage compared to the flat version. This reduction in storage usage highlights the efficiency of our proposed method.
Since works \cite{zhu2023sni, zhu2024semgauss, li2024sgs} report mIoU only on a subset of classes, making direct comparison unfair, we follow the same evaluation protocol as \cite{li2024sgs}, which computes mIoU using only the visible semantic classes in each frame. Our method achieves an mIoU of $96.79\%$ with a storage usage of $910.50$ MB, demonstrating on-par semantic rendering performance with state-of-the-art methods \cite{zhu2024semgauss}, while maintaining efficient storage usage.
Meanwhile, given that \cite{zhu2024semgauss} benefits significantly from a large foundation model pre-trained on much larger and more diverse datasets, our results demonstrate that comparable performance can be achieved without relying on such heavy models.

\begin{figure*}[!th]		
    \centering  
    \includegraphics[width=0.9\textwidth, trim=45mm 35mm 70mm 20mm, clip]{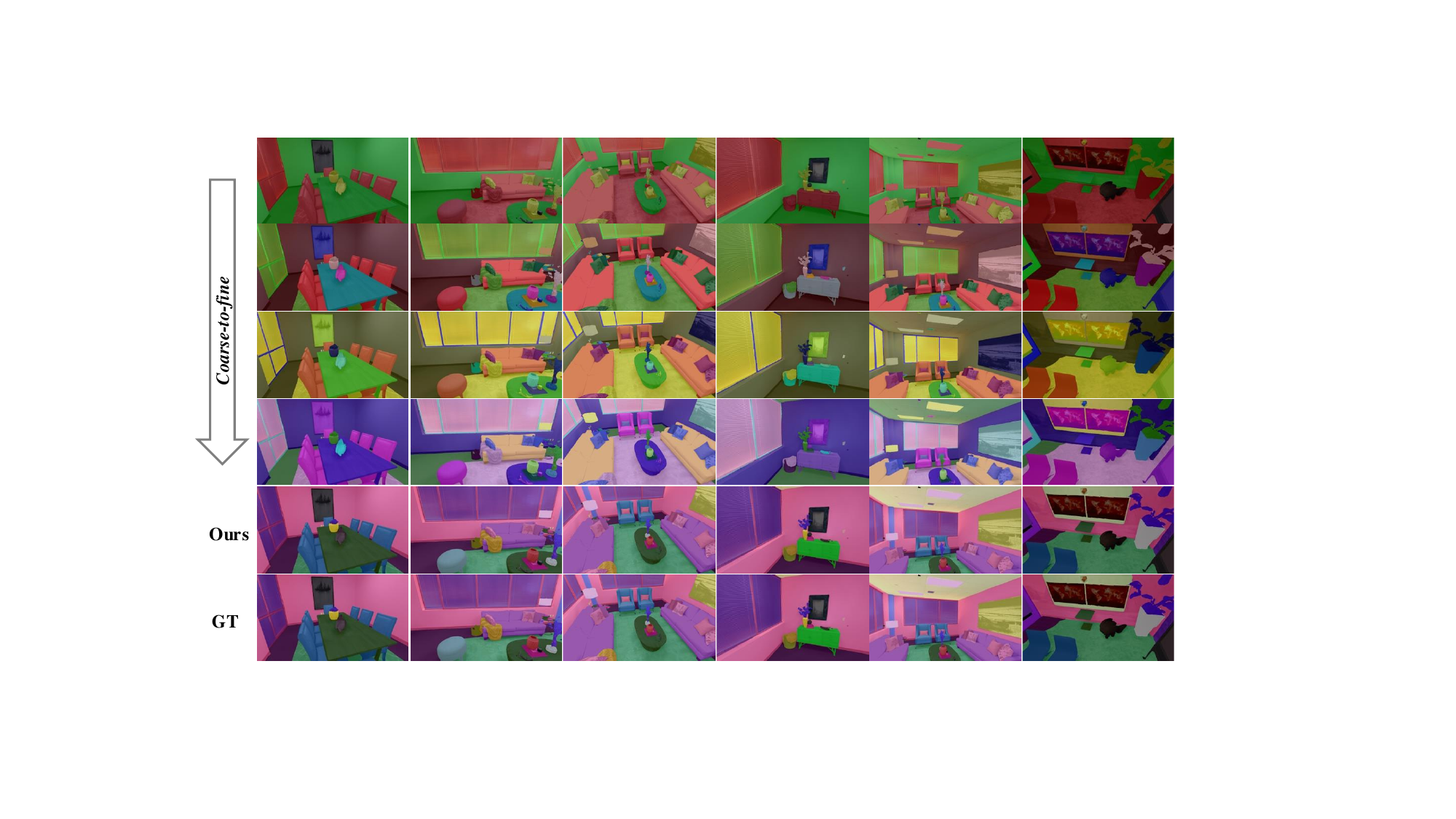} \\
    \caption{Visualization of our semantic rendering performance on the Replica \cite{straub2019replica} dataset. \textbf{The first four rows} demonstrate rendered semantic segmentation in a coarse-to-fine manner. \textbf{The fifth row} exhibits the finest semantic rendering, equivalent to the flat representation with $102$ original semantic classes from the Replica dataset. \textbf{The last row} visualizes the semantic ground truth for comparison.}  
    \label{fig:exp_sem_render} 
\end{figure*}

\Fig{fig:exp_sem_render} presents our qualitative results on the Replica dataset.  
We construct a five-level tree to hierarchically encode semantic classes, with the first five rows of \Fig{fig:exp_sem_render} illustrate the transition from level-0 to level-4, progressively refining the semantic understanding from coarse to detailed.
At the coarsest level (level-0), shown in the first row, segmentation is categorized into three broad classes: \textit{Small items}, \textit{Medium items}, \textit{Large items}.
In contrast, the finest level encompasses all 102 original semantic classes.
For example, the hierarchical understanding of the semantic label \textit{'Stool'} progresses from $\textit{Medium items} \rightarrow \textit{Seating furniture} \rightarrow \textit{Public seating} \rightarrow \textit{Flat} \rightarrow \textit{Stool}$, as demonstrated in the second column.
From \Fig{fig:exp_sem_render}, we can see that our method accurately renders semantics at each level, enabling a structured coarse-to-fine understanding of the entire scene.

Overall, our method demonstrates superior performance in semantic understanding while significantly reducing storage requirements and training time, benefiting from the proposed hierarchical semantic representation.

\begin{table}[!tb]
\centering
\setlength{\tabcolsep}{1.2pt}
\caption{Semantic performance mIoU (\%) and Parameter usage (MB) on Replica. Results are highlighted as \colorfirst, \colorsecond.} 
\label{tab:replica_semantic_segmentation}
\begin{tabular}{c|l|ccccccc}
\hline
\toprule 
& \textbf{Methods} & \textbf{Avg.} & \textbf{R0} & \textbf{R1} & \textbf{R2} & \textbf{Of0} \\ 
\hline
\multirow{8}{*}{\makecell{\textbf{mIoU (\%)} \\ total 102 classes}} 
& NIDS-SLAM \cite{haghighi2023neural} & 82.37 & 82.45 & 84.08 & 76.99 & 85.94 \\ 
& DNS-SLAM \cite{li2023dns} & 84.77 & 88.32 & 84.90 & 81.20 & 84.66 \\ 
& \oldMethodName{} \cite{li2024hi} & 76.44 & 76.62 & 78.31 & 80.39 & 70.43 \\
& \textbf{\MethodName{}} (flat) & \bestcolor{90.35} & \bestcolor{91.21} & \bestcolor{90.62} & 89.11 & \secondcolor{90.45} \\ 
& \textbf{\MethodName{}} (one-hot) & \secondcolor{89.41} & \secondcolor{86.38} & \secondcolor{89.26} & \bestcolor{91.55} & \bestcolor{90.46} \\ 
& \textbf{\MethodName{}} (binary) & 81.27 & 82.77 & 73.87 & \secondcolor{89.64} & 78.80 \\ 
\hline
\multirow{5}{*}{\makecell{\textbf{mIoU (\%)} \\ subset classes}}
 & SNI-SLAM \cite{zhu2023sni} & 87.41 & 88.42 & 87.43 & 86.16 & 87.63 \\
 & SemGauss-SLAM \cite{zhu2024semgauss} & \secondcolor{96.34} & \secondcolor{96.30} & \secondcolor{95.82} & \secondcolor{96.51} & \bestcolor{96.72} \\ 
 & SGS-SLAM \cite{li2024sgs} & 92.72 & 92.95 & 92.91 & 92.10 & 92.90 \\ 
 & \textbf{\MethodName{}} (one-hot)$^{\dag}$ &  \bestcolor{96.79} & \bestcolor{96.93} & \bestcolor{96.78} & \bestcolor{96.92} & \secondcolor{96.53} \\ 
 & \textbf{\MethodName{}} (binary)$^{\dag}$ & 95.26 & 96.21 & 92.62 & 96.34 & 95.85 \\

\hline
\multirow{4}{*}{\textbf{Param (MB)}} 
& \oldMethodName{} \cite{li2024hi} & \secondcolor{910.50} & \secondcolor{793} & \secondcolor{1126} & \secondcolor{843} & \secondcolor{880} \\ 
& \textbf{\MethodName{}} (flat) & 2662.25 & 2355 & 3072 & 2560 & 2662\\  
& \textbf{\MethodName{}} (one-hot) & 927.50 & 814 & \secondcolor{1126} & 867 & 903 \\ 
& \textbf{\MethodName{}} (binary) & \bestcolor{591.75} & \bestcolor{528} & \bestcolor{690} & \bestcolor{563} & \bestcolor{586} \\ 

\hline
\end{tabular}
\begin{tablenotes} 
            \footnotesize
            \itshape
            \raggedright 
            \item \textbf{Ours$^{\dag}$} represents our method with a hierarchical representation, evaluated on a subset of semantic classes, consistent with \cite{li2024sgs}.
        \end{tablenotes}
\end{table}

\begin{table}[!thb]
    \centering
    \caption{Monocular quantitative results on the TUM RGB-D dataset. Best results are highlighted as \colorfirst, \colorsecond.} 
    \setlength{\tabcolsep}{1.0pt} 
    \resizebox{1.0\linewidth}{!}{ 
    \begin{tabular}{cc|cccccccc}
    \hline
    \toprule 
	\multicolumn{2}{c}{Sequences} & \multicolumn{4}{c}{fr1-desk} & \multicolumn{4}{c}{fr2-xyz} \\
	\cmidrule(lr){1-2} \cmidrule(lr){3-6} \cmidrule(lr){7-10}
	{\scriptsize Loop} & {\scriptsize Method} & {\scriptsize RMSE (cm) $\downarrow$} & {\scriptsize PSNR $\uparrow$} & {\scriptsize SSIM $\uparrow$} & {\scriptsize LPIPS $\downarrow$} & {\scriptsize RMSE (cm) $\downarrow$} & {\scriptsize PSNR $\uparrow$} & {\scriptsize SSIM $\uparrow$} & {\scriptsize LPIPS $\downarrow$} \\
    \midrule      
    \multirow{3}{*}{{\rotatebox{90}{w/}}} & DROID-SLAM \cite{teed2021droid} & 1.80 & - & - & - & 0.50 & - & - & - \\
    & ORB-SLAM2 \cite{mur2017orb} & 2.00 & - & - & - & 0.60 & - & - & - \\
    & ORB-SLAM3 \cite{campos2021orb} & 1.89 & - & - & - & 0.29 & - & - & - \\
    & Photo-SLAM \cite{huang2024photo} & 1.54 & 20.97 & 0.743 & 0.228 & 0.98 & 21.07 & 0.726 & 0.166 \\
    \hdashline
    \multirow{3}{*}{{\rotatebox{90}{w/o}}} & DSO \cite{engel2017direct} & 22.4 & - & - & - & \bestcolor{1.10} & - & - & -  \\
    & MonoGS \cite{matsuki2024gaussian} & \bestcolor{4.15} & \secondcolor{21.02} & \secondcolor{0.703} & \secondcolor{0.362} & 4.79 & \secondcolor{22.21} & \secondcolor{0.728} & \secondcolor{0.288}  \\
    & \textbf{\MethodName{}} & \secondcolor{6.20} & \bestcolor{21.91} & \bestcolor{0.836} & \bestcolor{0.269} & \secondcolor{3.22} & \bestcolor{24.67} & \bestcolor{0.946} & \bestcolor{0.110}  \\ 
    \bottomrule
    \end{tabular}
    }
    \label{tab:tum_mono}
\end{table}

\begin{figure}[!t]		
    \centering  
    \includegraphics[width=0.5\textwidth, trim=0mm 67mm 0mm 0mm, clip]{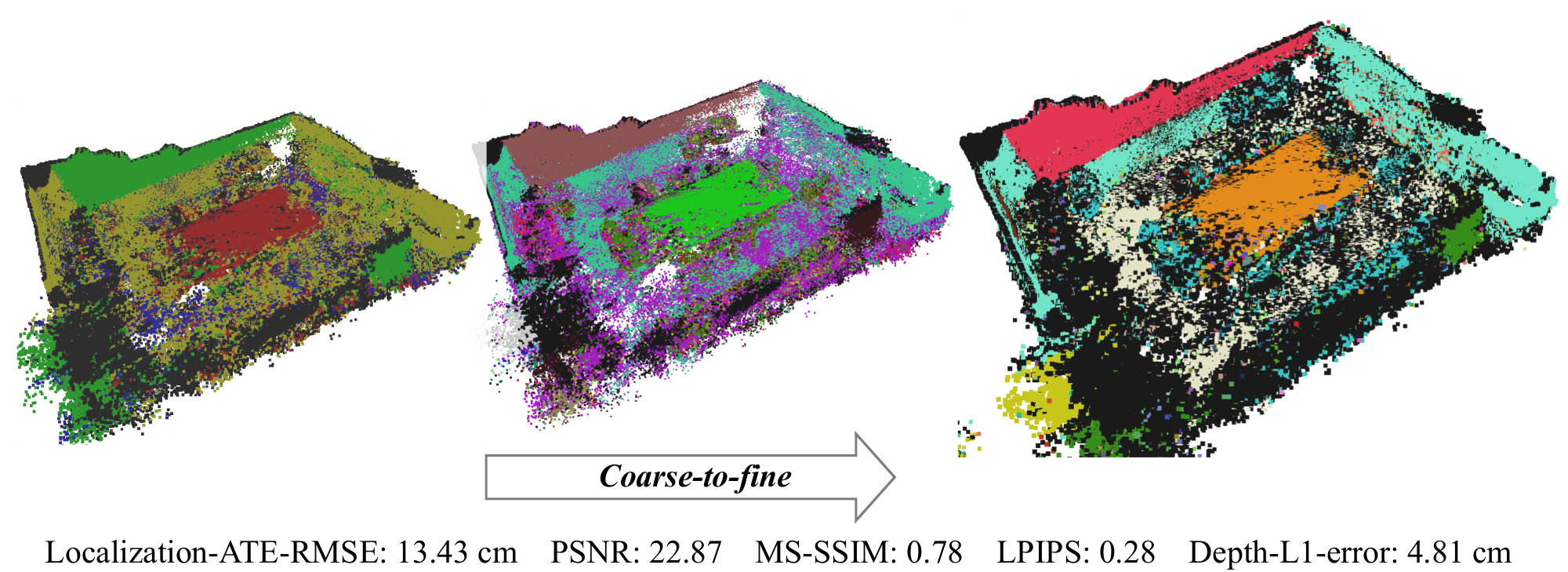} \\
    \caption{Visualization of the established semantic 3D map across multiple levels, illustrating a coarse-to-fine semantic understanding of the complex scene. The bottom of the figure presents localization, rendering, and depth performance, offering a comprehensive overview of \MethodName{}'s effectiveness and demonstrating the scaling-up capability of our proposed method.} 
    \label{fig:scannet_scaleup}  
    \vspace{-15pt}
\end{figure}

\subsection{Scaling up capability}

To showcase the scaling-up capability of \MethodName{}, we evaluate it on the real-world complex dataset, ScanNet \cite{Dai2017scannet}, which contains up to 550 distinct semantic classes. 
In contrast to Replica \cite{straub2019replica}, where the semantic ground truth is synthesized from a global world model and can be considered ideal, the semantic annotations in ScanNet are significantly noisier, containing considerable noise and incorrect or unclear class boundaries.
Moreover, the dataset contains noisy depth sensor inputs and blurred color images, making semantic understanding particularly challenging in this scenes.

Using the flat semantic representation cannot even run successfully due to memory limitations. 
In contrast, we establish the hierarchical tree with the assistance of LLMs and 3D Generative Models, as introduced in Section-III, which guides the compaction of the coding from the original 550 semantic classes to 73 semantic symbolic codings, resulting in over 7 times reduction in coding usage. As visualized in \Fig{fig:scannet_scaleup}, our estimated 3D global semantic map at different levels demonstrates a coarse-to-fine semantic understanding, showcasing our method's scaling-up capability in handling real-world complex scenes.

\subsection{Monocular performance}

We present our monocular SLAM performance in \Tab{tab:tum_mono}.
For SLAM methods without loop closure, our approach achieves competitive results compared to state-of-the-art methods in the same setting.
For methods with loop closure, where well-executed loop mechanisms typically boost tracking accuracy and overall SLAM performance, our method still delivers better rendering accuracy compared to Gaussian-based SLAM systems~\cite{huang2024photo}.
With the assistance of the geometric prior provided by the feed-forward model, the joint optimization of both photometric loss and aligned depth loss contributes to improved rendering performance across the entire SLAM system.

\subsection{Ablation study}

In this section, we evaluate the SLAM performance under different hierarchical tree designs. The results are presented in \Tab{tab:ab_diff_tree}.
Specifically, the experiments are conducted using our baseline setup, which adopts a one-hot hierarchical semantic representation and a naïve decoder with a single Softmax layer, denoted as \textit{Baseline} in the table.
This setup ensures that any observed performance differences are solely attributable to the tree structure itself.
Each metric in \Tab{tab:ab_diff_tree} represents the average result over two runs on the first four sequences—Room0, Room1, Room2, and Office0—from the Replica dataset.
We generate two distinct 5-level hierarchical trees (tree1 and tree2) using our proposed tree construction framework, which integrates both LLM-generated semantics and 3D generative priors, and evaluate them within our semantic SLAM system.
For comparison, we also include our previous Hier-SLAM~\cite{li2024hi} (tree0), which relies solely on LLM-generated semantics.
As shown in the table, all configurations yield strong performance, which we attribute to the quality of the tree structures produced by the joint LLM and 3D prior-based generation process.
In addition, the fully optimized hierarchical semantic loss contributes to the robustness of SLAM performance across tree variants.
Among the compared configurations, tree2 achieves the highest semantic mIoU, the best depth reconstruction accuracy, and comparable rendering quality (in PSNR, SSIM, and LPIPS) relative to the alternatives.
Therefore, we adopt tree2 as the default hierarchical structure in our semantic SLAM framework.

\begin{table}[!tb]
    \centering
    \caption{Performance of different trees on Replica. BEST RESULTS ARE HIGHLIGHTED AS \textbf{FIRST}.}
    \vspace{3pt}
    \renewcommand{\arraystretch}{1.1}
    \setlength{\tabcolsep}{1pt}
    \label{tab:ab_diff_tree}
    \begin{tabular}{lcccccc}
        \toprule
         Methods & \textbf{mIou}$\uparrow$ & \textbf{ATE RMSE}$\downarrow$ & \textbf{Depth}$\downarrow$ & \textbf{PSNR}$\uparrow$ & \textbf{MS-SSIM}$\uparrow$ & \textbf{LPIPS}$\downarrow$ \\
        \midrule
        Baseline (tree0)  & 75.517 & 0.292 & 0.416 & \textbf{35.885} & \textbf{0.982} & \textbf{0.057} \\
        Baseline (tree1)  & 75.061 & \textbf{0.280} & 0.413 & 35.792 & \textbf{0.982} & \textbf{0.057} \\ 
        Baseline (tree2)  & \textbf{76.754} & 0.302 & \textbf{0.408} & 35.825 & \textbf{0.982} & 0.058 \\ 
        \bottomrule
    \end{tabular}
\end{table}

We also evaluate our proposed one-hot and binary semantic embeddings in previous experiments, including camera tracking performance in \Tab{tab:exp_pose_replica} and \Tab{tab:exp_pose_scannet}, reconstruction accuracy in \Tab{tab:exp_depth}, rendering quality in \Tab{tab:exp_render_replica}, as well as runtime and semantic understanding in \Tab{tab:runtime} and \Tab{tab:replica_semantic_segmentation}, respectively.
For camera tracking, mapping, and rendering performance, both embeddings exhibit similar results.
However, for runtime and semantic understanding, the two embeddings show differences. Specifically, our one-hot representation achieves better semantic understanding (higher mIoU in \Tab{tab:replica_semantic_segmentation}), while the binary embedding shows an 8\% drop in mIoU on the full semantic class evaluation and less than 1\% drop on the subset evaluation, but provides faster operation and significantly reduced storage—approximately 36\% savings, as illustrated in \Tab{tab:runtime} and \Tab{tab:replica_semantic_segmentation}.
As introduced in Section III, our proposed semantic embeddings provide two types of compactness: the one-hot representation effectively compresses semantic information, achieving faster training and lower memory usage, while the binary embedding further reduces storage and computation time at the cost of a small decrease in semantic accuracy (which remains minimal on subset evaluations).
For tracking, mapping, and rendering, the performance difference between the two embeddings is negligible, as both integrate semantic information into the SLAM process with interpretable compactness derived from our carefully designed hierarchical tree representation.

%% file: text/5_conclusion.tex
\section{Conclusion}

We introduced \MethodName{}, a neuro-symbolic semantic 3D Gaussian Splatting SLAM system that supports both RGB-D and monocular inputs, incorporating a novel hierarchical categorical representation.  
Specifically, we proposed a compact and generalized hierarchical representation that encodes both semantic meaning and geometric attributes (i.e., size and shape), leveraging the capabilities of LLMs and 3D generative models.  
To enhance global semantic understanding, we introduced a novel hierarchical semantic loss.  
Furthermore, we extended the system from an RGB-D-only SLAM to support monocular input, utilizing geometric priors derived from a feed-forward SLAM approach.  
Experimental results demonstrate that \MethodName{} achieves superior or on-par performance with state-of-the-art NeRF-based and Gaussian-based SLAM methods in terms of tracking, mapping, and semantic understanding accuracy, while significantly reducing storage requirements and runtime.
These advances establish \MethodName{} as a robust solution for semantic 3D understanding across diverse environments.




 








%% file: text/6_appendix.tex
\clearpage
\newpage
\setcounter{section}{0}
\setcounter{figure}{0}

\begin{center}
    \textbf{\large APPENDIX}
\end{center}

\section{Implementation Details for Tree Generation}
\normalfont

During the hierarchical tree generation process, we utilize LLMs to group flat semantic classes at different levels based on \textbf{physical size}, \textbf{semantic function}, and to \textbf{summarize and refine geometric grouping}, ensuring balanced, meaningful, and descriptive outputs.
After generating each tree level, we further employ LLMs as \textbf{validators} to examine the results at each level, ensuring that all input categories are fully included with no omissions and that no extra categories are mistakenly introduced.

\noindent\textbf{Size Grouping: }
The prompt used for the initial grouping based on object physical sizes is illustrated as follows:

\begin{prompt}[title={Code \thetcbcounter: Prompt for Size Grouping}] \label{promt:size_cluster}
You are a smart assistant tasked with dividing the following items into meaningful groups based on their properties. The key requirements are:\\
\textbf{Balance:} The number of items in each group should be as evenly distributed as possible. A difference of 1 item between groups is acceptable, but larger differences should be avoided. \\
\textbf{Meaningfulness:} The groups must be meaningful, based on the items' inherent characteristics. Grouping should make logical sense to a human observer. \\
\textbf{Descriptive Group Names:} Each group must have a clear and descriptive name that reflects its characteristics. \\
\textbf{Goal:} Cluster items into size groups, based on their typical physical size in scenes. The groups must strictly be by sizes, for example: \texttt{small, small medium, medium, large, extra large}, etc.\\
\textbf{Items:} \texttt{\{classes input\}}

Ensure groups are meaningful and provide descriptive group names. Output must follow this JSON format:
\begin{verbatim}
{
    "<GROUP_1>": ["<ITEM_1>", ...], 
    "<GROUP_2>": ["<ITEM_2>", ...], 
    ...
}
\end{verbatim}
\end{prompt}

\noindent\textbf{Function Grouping: }
The prompt adopted for the subsequent semantic function grouping is as follows:

\begin{prompt}[title={Code \thetcbcounter: Prompt for Function Grouping}] \label{promt:func_cluster}
You are a smart assistant tasked with dividing the following items into meaningful groups based on their properties. The key requirements are:\\
\textbf{Balance:} The number of items in each group should be as evenly distributed as possible. A difference of 1 item between groups is acceptable, but larger differences should be avoided. \\
\textbf{Meaningfulness:} The groups must be meaningful, based on the items' inherent characteristics. Grouping should make logical sense to a human observer. \\
\textbf{Descriptive Group Names:} Each group must have a clear and descriptive name that reflects its characteristics. \\
\textbf{Goal:} Cluster items in the '{size}' size group into functionality-based groups.
The name of the clusters should not be too specific, it could be as general like \texttt{storage, furniture}, etc.
Ensure that items in each group serve similar purposes or have similar functionalities.\\
\textbf{Items:} \texttt{\{classes input\}}

Ensure groups are meaningful and provide descriptive group names. Output must follow this JSON format:
\begin{verbatim}
{
    "<GROUP_1>": ["<ITEM_1>", ...], 
    "<GROUP_2>": ["<ITEM_2>", ...], 
    ...
}
\end{verbatim}
\end{prompt}

\noindent\textbf{Summarize and Refine for Geometric Grouping: }
After grouping with the 3D Generative Model, we employ LLMs to summarize the clustering results with descriptive labels and balance the nodes within each group. The prompt for LLMs is illustrated as follows:

\begin{prompt}[title={Code \thetcbcounter: Prompt for Geometric Grouping Results}] \label{promt:geom_cluster}
The following groups of indoor scene items have been clustered based on their shape: \texttt{\{formatted\_clusters\}}\\
\textbf{Instructions:} 
\textbf{Balance:} Ensure the groups are as evenly distributed as possible. A difference of 1 item between groups is acceptable. If needed, move a small number of items from one group to another to achieve balance, feel free to even remove a group if it has only one member (just dont leave any groups empty), ensuring that the groups remain meaningful. \\
\textbf{Meaningful Naming:} After balancing the groups, assign a descriptive and meaningful name to each group, based on the shared shape characteristic. The name should clearly reflect the shape or geometric property of the items in that group. Always prefer singular shapes names like box, rectangle, soft, flat, etc. \\
\textbf{No Duplications:} make sure to not repeat any class members, the group names are fine but not the groups' members. \\
The group names must strictly by shapes and DO NOT leave any group empty, you could remove it if its empty.\\
The output must be the same JSON format as below:\\
\begin{verbatim}
{
    "<GROUP_1>": ["<ITEM_1>", ...], 
    "<GROUP_2>": ["<ITEM_2>", ...], 
    ...
}
\end{verbatim}
\end{prompt}

\noindent\textbf{LLM Validators: }
The prompt used for LLM validator is as following:

\begin{prompt}[title={Code \thetcbcounter: Prompt for LLM validator}] \label{promt:llm_validator}
 The following groups of items in a scene have been clustered based on their shape: \texttt{\{formatted\_clusters\}}\\
\textbf{Instructions:} 
\textbf{Balance:} Ensure the groups are as evenly distributed as possible. A difference of 1 item between groups is acceptable. If needed, move a small number of items from one group to another to achieve balance, remove a group if it has only one member, ensuring that the groups remain meaningful. \\
\textbf{Meaningful Naming:} After balancing the groups, assign a descriptive and meaningful name to each group, based on the shared shape characteristic. The name should clearly reflect the shape or geometric property of the items in that group. Always prefer singular shapes names like box, rectangle, soft, flat, etc. \\
\textbf{New Group:} If it is necessary to create a new group, feel free to do so, DO NOT create any non-original items. \\
\textbf{No Duplications:} make sure to not repeat any class members, the group names are fine but not the groups' members. \\
The output must be the same JSON format as below:\\
\begin{verbatim}
{
    "<GROUP_1>": ["<ITEM_1>", ...], 
    "<GROUP_2>": ["<ITEM_2>", ...], 
    ...
}
\end{verbatim}
\end{prompt}


\section{Generated tree demonstration}

\begin{figure*}[!thb]		
    \centering  
\includegraphics[width=1.0\textwidth, trim=30mm 30mm 30mm 30mm, clip]{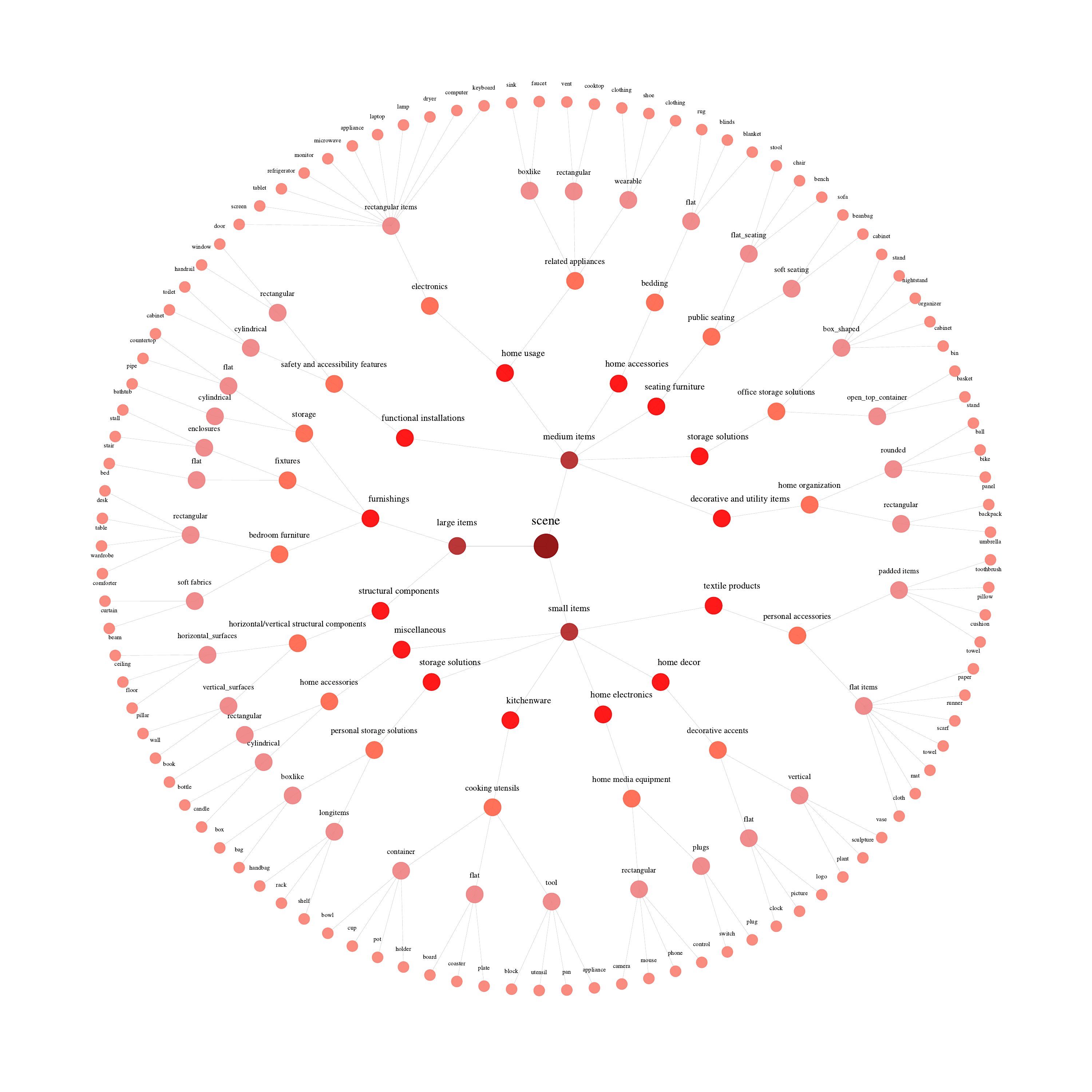} \\
    \caption{Visualization of the generated hierarchical tree on the Replica dataset~\cite{straub2019replica}. Different colors represent different tree depths. Each node corresponds to a symbolic concept, with label names shown next to the nodes.} 
    \label{fig:tree_replica} 
\end{figure*}

We demonstrate the generated hierarchical tree for the Replica dataset~\cite{straub2019replica} in \Fig{fig:tree_replica}.
Based on the semantic categories in Replica, we leverage the capabilities of large language models and 3D generative models to construct a hierarchical tree.
This tree represents semantic information as symbolic nodes (illustrated as colored nodes), which are used to produce compact semantic embeddings for each Gaussian primitive. These embeddings are then learned end-to-end across multiple views during SLAM operation.

